\documentclass[10pt, a4paper]{article}

\usepackage[final]{lrec2026} 

\usepackage{setspace}

\usepackage{times}
\usepackage{latexsym}

\usepackage[T1]{fontenc}

\usepackage[utf8]{inputenc}

\usepackage[scale=0.99]{inconsolata}

\graphicspath{{.}}

\usepackage{booktabs}
\usepackage{tabularx}
\usepackage{longtable}

\usepackage{caption}

\usepackage{amsmath}

\usepackage{enumitem}

\usepackage{xcolor}

\usepackage{float}

\usepackage{listings}
\lstdefinestyle{codestyle}{
    backgroundcolor=\color{BgColor},   
    commentstyle=\color{CodeGreen},
    keywordstyle=\color{CustomBlue},
    numberstyle=\tiny\color{CustomDarkGray},
    stringstyle=\color{CustomAccentGreen},
    basicstyle=\fontfamily{zi4}\footnotesize,
    moredelim=**[is][\color{CustomDarkBlue}]{@}{@},
    breakatwhitespace=true,         
    breaklines=true,
    breakindent=0pt,
    breakautoindent=false,
    captionpos=b,                    
    keepspaces=true,                 
    numbers=left,                    
    numbersep=5pt,                  
    showspaces=false,                
    showstringspaces=false,
    showtabs=false,                  
    tabsize=2,
    escapeinside={<@}{@>}
}

\usepackage{algorithm}
\usepackage[noend]{algpseudocode}
\newcommand\CONDITION[2]%
{\begin{tabular}[t]{@{}l@{}}
		#1 #2
 \end{tabular}%
}
\makeatletter
\renewcommand{\ALG@beginalgorithmic}{\small}
\renewcommand{\ALG@name}{Algorithm}
\makeatother
\algdef{SE}[WHILE]{While}{EndWhile}[1]%
{\algorithmicwhile\ \CONDITION{#1}{\ \algorithmicdo}}%
{\algorithmicend\ \algorithmicwhile}

\algdef{SE}[FOR]{For}{EndFor}[1]%
{\algorithmicfor\ \CONDITION{#1}{\ \algorithmicdo}}%
{\algorithmicend\ \algorithmicfor}

\algdef{S}[FOR]{ForAll}[1]%
{\algorithmicforall\ \CONDITION{#1}{\ \algorithmicdo}}

\algdef{SE}[REPEAT]{Repeat}{Until}{\algorithmicrepeat}[1]%
{\algorithmicuntil\ \CONDITION{#1}{}}

\algdef{SE}[IF]{If}{EndIf}[1]%
{\algorithmicif\ \CONDITION{#1}{\ \algorithmicthen}}%
{\algorithmicend\ \algorithmicif}%

\algdef{C}[IF]{IF}{ElsIf}[1]%
{\algorithmicelse\ \algorithmicif\ \CONDITION{#1}{\ \algorithmicthen}}

\newenvironment{prompt}%
{\begin{quote}\parindent0pt\ttfamily\color{CustomDarkBlue}}
{\end{quote}}

\newcommand{\monospace}[1]{\texttt{\textcolor{CustomDarkBlue}{#1}}}

\usepackage[oneside]{rotating}
\usepackage{multirow}

\newcommand{\q}[1]{``#1''}  

\definecolor{CustomBlue}{HTML}{0065BD}
\definecolor{CustomAccentBlue}{HTML}{64A0C8}
\definecolor{CustomDarkBlue}{HTML}{19127B}
\definecolor{CustomDarkGray}{HTML}{333333}
\definecolor{CustomAccentGreen}{HTML}{A2AD00}
\definecolor{CodeGreen}{rgb}{0,0.6,0}
\definecolor{CodeGray}{rgb}{0.5,0.5,0.5}
\definecolor{CodePurple}{rgb}{0.58,0,0.82}
\definecolor{BgColor}{HTML}{F6F8FA}
\definecolor{SoftOrange}{RGB}{255, 230, 230}
\definecolor{SoftGreen}{RGB}{228, 255, 226}

\newcommand{\anneq}[1]{\sethlcolor{SoftGreen}\hl{\{\{#1\}\}}}
\newcommand{\anndiss}[1]{\sethlcolor{SoftOrange}\hl{\{\{#1\}\}}}

\let\oldfootnote\footnote
\renewcommand{\footnote}[1]{\oldfootnote{\hspace{0.1mm} #1}}

\title{Explainable Semantic Textual Similarity \\ via Dissimilar Span Detection}

\name{Diego Miguel Lozano\textsuperscript{1, \textdagger}\thanks{\textsuperscript{\textdagger} Currently affiliated to ELLIS Alicante.}, Daryna Dementieva\textsuperscript{1,2}, Alexander Fraser\textsuperscript{1,2}} 

\address{\textsuperscript{1}School of Computation, Information and Technology \\
         Technical University of Munich \\
         \textsuperscript{2}Munich Center for Machine Learning (MCML) \\
         \texttt{\href{mailto:diego@ellisalicante.org}{diego@ellisalicante.org},
         \{\href{mailto:daryna.dementieva@tum.de}{daryna.dementieva}, \href{mailto:alexander.fraser@tum.de}{alexander.fraser}\}@tum.de}\\}

\abstract{
Semantic Textual Similarity (STS) is a crucial component of many Natural Language Processing (NLP) applications. However, existing approaches typically reduce semantic nuances to a single score, limiting interpretability. To address this, we introduce the task of Dissimilar Span Detection (DSD), which aims to identify semantically differing spans between pairs of texts. This can help users understand which particular words or tokens negatively affect the similarity score, or be used to improve performance in STS-dependent downstream tasks. Furthermore, we release a new dataset suitable for the task, the Span Similarity Dataset (SSD), developed through a semi-automated pipeline combining large language models (LLMs) with human verification. We propose and evaluate different baseline methods for DSD, both unsupervised---based on LIME, SHAP, LLMs, and our own method---as well as an additional supervised approach. While LLMs and supervised models achieve the highest performance, overall results remain low, highlighting the complexity of the task. Finally, we set up an additional experiment that shows how DSD can lead to increased performance in the specific task of paraphrase detection.
 \\ \newline }

\begin{document}

\maketitleabstract

\section{Introduction}

Semantic Textual Similarity (STS) is a fundamental concept in Natural Language Processing (NLP), being present in a myriad of tasks. For example, STS is the cornerstone of paraphrase identification \citep{zhou2022paraphrase}, it is widely used for text classification and clustering tasks \citep{minaee2021deep}, it lays the foundation for popular evaluation metrics, such as BERTScore \citep{zhang2020bertscore} and, more recently, it has enabled the development of Dense Passage Retrieval (DPR) \citep{karpukhin2020dense}, as well as some other highly active areas of research such as Retrieval-Augmented Generation (RAG) \citep{lewis2021retrievalaugmented}, which has also attracted great attention from the industry.

However, most frequently, a single similarity score is reported (e.g., cosine similarity), and condensing all the semantic nuances into this score hinders interpretability. Popular techniques in the field of Explainable AI (XAI) revolve around natural language explanations (also called rationales, \citealp{gurrapu2023}), or visualizations. These also fit well within the context of STS, and in the present work, we focus on providing explanations through textual highlighting. More concretely, we want to identify span pairs that differ in meaning, a task we call Dissimilar Span Detection (DSD).

\begin{figure*}[t]
  \includegraphics[width=\textwidth]{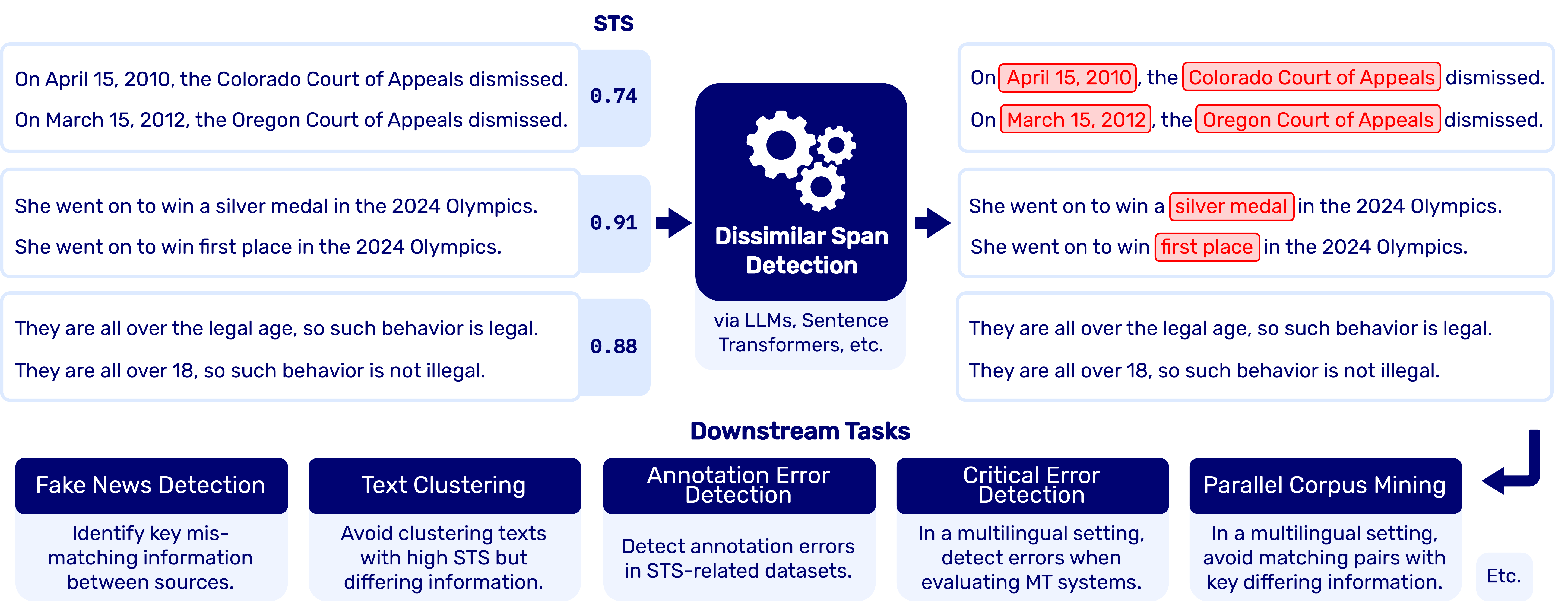}
  \caption{Examples of the \textit{Dissimilar Span Detection} task, i.e., given a pair of texts, identify which spans differ semantically. Here, we show two pairs that contain dissimilar spans and one pair that does not. Cosine similarities are calculated with the Sentence Transformer model \texttt{all-MiniLM-L6-v2}. Note that pairs containing dissimilar spans might still yield a high semantic textual similarity, sometimes even higher than sentences that are equivalent in meaning. Therefore, relying on STS alone could potentially lead to erroneous decisions in downstream tasks.}
  \label{fig:dissimilar-span-detection}
\end{figure*}

More formally, given two sequences of word or sub-word tokens $X = (x_1, x_2, ..., x_n)$ and $Y = (y_1, y_2, ..., y_m)$, let $s_j := ((x_a, x_{a+c}), (y_b, y_{b+d}))$ be a token span pair coming from $X$ and $Y$, with $1 \leq a < n$, $1 \leq b < m$, $1 \leq c \leq n - a$ and $1 \leq d \leq m - b$, the goal is to find all non-overlapping span pairs $s_j$ that, having a common semantic function, differ in meaning. Note that a text pair might also contain no such spans.

As seen in \autoref{fig:dissimilar-span-detection}, the annotations can directly be shown to users in addition to the overall textual similarity, as well as be used to support other downstream tasks such as: Fake News Detection, in order to detect key mismatching information; Annotation Error Detection \citep{klie-etal-2023-annotation}, e.g., to detect errors in STS-related datasets; Text Clustering or Paraphrase Identification, to prevent clustering together or flagging as paraphrase texts that contain small but critical differences; finally, when taken to a multilingual setting, the task of DSD is very similar to that of Critical Error Detection \citep{zerva-etal-2022-findings}, i.e., predicting whether a translation contains a critical error or not, and could also assist in Parallel Corpus Mining, to avoid matching similar texts that contain key discrepancies.

Our main contributions are:

\begin{itemize}[itemsep=0.1pt,topsep=1pt]
    \item We introduce the task of Dissimilar Span Detection (DSD): a method that provides interpretability in the context of STS which involves identifying span pairs that differ in meaning. This method can also improve performance in various downstream, STS-dependent tasks.
    \item We propose a semi-automatic pipeline to create suitable training and evaluation data for the task. A dataset produced in such way, the so-called Span Similarity Dataset (SSD), is publicly released, along with a comprehensive evaluation setup. The dataset's correctness is validated through the manual annotation of a subset by 6 different annotators, which we compare against the semi-automatically annotated data.
    \item We evaluate a number of baseline methods to tackle the task. In particular, two unsupervised, model-agnostic methods based on LIME \citep{ribeiro2016why} and SHAP \citep{NIPS2017_7062} are evaluated. Inspired by the latter, we propose a further method that yields better results. We also consider state-of-the-art LLMs, which perform best overall, and evaluate different supervised models fine-tuned on the task, obtaining the best compromise between size and performance. Nevertheless, results stay generally low, highlighting the difficulty of DSD even in our simplified setting. Finally, we apply DSD to the task of paraphrase detection, showing its viability to improve performance in downstream tasks.
\end{itemize}

The source code and the dataset are publicly available at \url{https://dmlls.github.io/dissimilar-span-detection}, licensed under GPL v3.0 and CC BY-SA 4.0, respectively.

\begin{table*}[t]
  \centering
  \begin{tabularx}{\textwidth}{XXcc}
    \toprule
      \small{\textbf{Sentence 1}} & \small{\textbf{Sentence 2}} & \small\begin{tabular}{@{}c@{}} \textbf{Span} \\ \textbf{Similarity} \end{tabular} & \small\begin{tabular}{@{}c@{}} \textbf{Sentence} \\ \textbf{Similarity} \end{tabular} \\
    \midrule

      \footnotesize\begin{tabular}{@{}l@{}} It only utilizes \anndiss{a few refineries} \\ in the Northeast. \end{tabular} & \footnotesize\begin{tabular}{@{}l@{}} It only utilizes \anndiss{numerous facilities} \\ in the Northeast. \end{tabular}  & \footnotesize{0} & \footnotesize{0} \\[4.5mm]

      \footnotesize\begin{tabular}{@{}l@{}} It was \anndiss{restored} in the \anneq{1980s}. \end{tabular} & \footnotesize\begin{tabular}{@{}l@{}} It was \anndiss{destroyed} in the \anneq{eighties}. \end{tabular}  & \footnotesize{0,1} & \footnotesize{0} \\[3mm]

      \footnotesize\begin{tabular}{@{}l@{}} Thank you for \anneq{wasting my money}. \end{tabular} & \footnotesize\begin{tabular}{@{}l@{}} Thank you for \anneq{misusing my funds}. \end{tabular}  & \footnotesize{1} & \footnotesize{1} \\[3mm]

      \footnotesize\begin{tabular}{@{}l@{}} There are depots at \anndiss{Quilpie} and \\ \anndiss{Roma}. \end{tabular} & \footnotesize\begin{tabular}{@{}l@{}} There are depots at \anndiss{Brisbane} and \\ \anndiss{Sydney}. \end{tabular}  & \footnotesize{0,0} & \footnotesize{0} \\
    \bottomrule
  \end{tabularx}
  \caption{Examples of shorter sentences from the SSD. Spans are denoted using double curly braces, and then annotated respectively with a 0, in case the span pair differs in meaning, or 1, if they are equivalent. If there exist several annotated spans, these values are separated by a comma. Finally, the sentence similarity column denotes whether the meaning of the sentence pair as a whole is equivalent in meaning (1) or not (0). We have highlighted similar (green) and dissimilar spans (red) for better readability.}\label{tab:ssd-example}
\end{table*}

\section{Related Work}

The task of annotating tokens or spans of text to provide explanations has already been worked on, albeit not extensively, in the context of Natural Language Inference (NLI). A popular dataset in this regard is e-SNLI \citeplanguageresource{NIPS2018_8163}, which extends the Stanford Natural Language Inference (SNLI) corpus \citeplanguageresource{bowman-etal-2015-large} by highlighting the words most relevant for a particular entailment label, and providing natural language explanations for these annotations.

\citet{thorne-etal-2019-generating} explored different methods, namely LIME and Anchor Explanations \citep{ribeiro2018anchors}, as well as their own method, based on Multiple Instance Learning (MIL), in order to generate token-level explanations that justify the NLI labels. A continuation in this direction is \citet{youngwoo2020explaining}, which expands previous work by annotating not only the tokens/spans relevant for the entailment prediction, but also classifying the detected spans into different categories.

Moving more concretely to the topic of explainability in the context of STS, \citeplanguageresource{Rus2012TheSC} released the SIMILAR corpus in an effort to improve the explainability of word-to-word similarity metrics. It contains 700 manually annotated sentences and considers different types of semantic relations between words.

Still working at the word level, \citet{malkiel2022interpreting} introduce an unsupervised technique named BTI (BERT Interpretations) that employs BERT-like models to explain similarities between paragraphs. In this case, individual, matching words that best explain the similarity between pairs of texts are identified. The selected word pairs can then be visualized by the user, helping to establish connections between the two texts.

One of the works most closely related to ours is \citet{Lopez_Gazpio_2017}, which deals with the concept of \textit{Interpretable Semantic Textual Similarity}, and is framed within the Task 2 of both SemEval-2015 \citep{semeval-2015} and SemEval-2016 \citep{semeval-2016}. The authors formalized the interpretability layer as \q{the alignment between pairs of segments\footnote{Segments are what we refer to as \textit{spans}.} across the two sentences, where the relation between the segments is labeled with a relation type and a similarity score.} Granular span labels are defined, such as: \texttt{EQUI}, for semantically \textit{equivalent} spans; \texttt{SIM}, for \textit{similar} spans; or \texttt{OPPO}, for spans with \textit{opposite} meanings. Each of these mutually exclusive labels is also accompanied by a number ranging from 1 to 5, indicating the level of similarity or equivalence. Apart from proposing different methods to solve the task, the authors also implemented a user-facing verbalization system. The dataset they used had been previously released for Task 2 of SemEval-2016 \citeplanguageresource{semeval-2016-task-2}, which we also use as an additional resource to evaluate the different baseline methods we consider.

However, to the best of our knowledge, there has been no prior research addressing the utility of detecting semantic \textit{dissimilarities} at the span level. This focus on dissimilarity can be useful not only to provide explanations to end users---an effective way to explain similarity, or the lack thereof, is by pointing out the dissimilarities---but also to improve the performance of systems in downstream tasks that rely on STS.

\vspace{0.2cm}
\section{Span Similarity Dataset (SSD)}

In this section, we discuss the motivation behind building a new dataset for the task and describe how it was constructed. We also conduct a dataset analysis, reporting several statistics about it.

\subsection{Dataset Motivation}

We initially considered NLI-related datasets like e-SNLI. However, pairs labeled as \textit{contradiction} (i.e., those that contain dissimilarities) are often so different that the entire sentence would need to be labeled as dissimilar.\footnote{In fact, the mean cosine similarity of all the \textit{contradiction} pairs in the e-SNLI dataset (considering the \textit{train}, \textit{dev}, and \textit{test} splits) is as low as 0.31, with a standard deviation of 0.20 (calculated using the Sentence Transformer model \texttt{all-mpnet-base-v2}).} Furthermore, the tasks of NLI and STS, although related, are fundamentally different. It is common for NLI datasets to contain sentences that may be logically related in ways that do not necessarily reflect their semantic similarity. As a result, employing these datasets for evaluating DSD would not have been reliable.

On the other hand, we also considered the aforementioned dataset from SemEval-2016, which is specifically crafted for interpretable STS. Unfortunately, it was still too restricted. Even though this dataset annotates different relations between spans, in our case we were only able to use those labeled as \textit{opposite} (\texttt{OPPO}), since all other categories include a mix of similar and dissimilar spans (or no dissimilar spans at all). For example, spans such as ``a woman'' and ``a man'', or ``17 civilians'' and ``10 police'' were annotated as \textit{similar}, when in our context these pairs are considered dissimilar. Consequently, out of 2,929 sentence pairs, there were only 109 relevant to our context.

Due to the scarcity of suitable data, we found it necessary to create our own dataset, the so-called Span Similarity Dataset (SSD). It currently contains 1,000 annotated samples that specifically target the task of DSD. As previously mentioned, it has been publicly released for the benefit of the NLP community.

\subsection{Dataset Construction}

The goal of our dataset is to provide a first, simplified DSD benchmark to assess where state-of-the-art models, such as LLMs, stand in relation to this task. Thus, we limit ourselves to English and stay at the sentence level.

The sentences of the SSD were sourced from a random subset of the samples from the CANNOT Dataset \citeplanguageresource{cannot-dataset}, which in turn is a compilation of data from different datasets targeting tasks closely related to DSD, such as fact-checking \citep{sathe-etal-2020-automated}, paraphrase detection \citep{vahtola-etal-2022-easy}, or negation identification \citep{truong-etal-2022-another}.

The main steps involved in the annotation were:

\begin{enumerate}[itemsep=0.05pt,topsep=1.2pt]
    \item Taking the first sentence and altering one or more spans of words, giving result to the second sentence. The modified spans could either be equivalent in meaning to the original one, or be semantically dissimilar.
    \item Enclosing each of the altered spans between span annotation markers. In our case, \texttt{\{\{} denotes the beginning of a span, and \texttt{\}\}} its end.
    \item Annotating each of the span pairs with either a 1, if they are equivalent in meaning, or 0 otherwise.
    \item Annotating whether the entire two sentences are semantically equivalent (1) or not (0).
\end{enumerate}

The annotation was performed in a semi-automatic way through the use of an LLM\footnote{ChatGPT \citep{chatgpt} with the GPT-3.5-Turbo backend was employed between November 2023 and April 2024.} via a manually engineered prompt, significantly reducing time and effort by allowing the model to replace spans and assign labels. Nevertheless, since the model was unable to consistently annotate the correct labels, we decided to perform the span labeling manually. Finally, each of the automatically generated spans was manually reviewed, and if needed, corrected, before being added to the dataset. For reference, annotating 100 samples this way takes between 1 and 2 hours. An example of the resultant samples can be seen in \autoref{tab:ssd-example}.

\subsection{Dataset Analysis}

The main statistics of the SSD are collected in \autoref{tab:sdd-statistics}. Furthermore, we perform POS tagging on the dataset, obtaining the most frequent POS labels enclosed within spans: nouns (38.12 \%), proper nouns (21.72 \%), verbs (16.44 \%), and adjectives (13.44 \%). As for grammatical dependencies, the most common are: objects of a preposition -- \textit{pobj} (23.03 \%), compound entities (15.72 \%), and adjectival modifiers -- \textit{amod} (10.69 \%).

\begin{table}
  \centering
  \begin{tabularx}{\columnwidth}{Xc}
    \toprule
    \small{\textbf{Attribute}} & \small{\textbf{Value}} \\
    \midrule
      \footnotesize{Sentence pairs} & \footnotesize{1,000} \\
      \footnotesize{Span pairs} & \footnotesize{1,296} \\
      \footnotesize{Dissimilar span pairs} & \footnotesize{648} \scriptsize{(50 \%)} \\
      \footnotesize{Equivalent span pairs} & \footnotesize{648} \scriptsize{(50 \%)} \\
      \footnotesize{Dissimilar sentence pairs} & \footnotesize{571} \scriptsize{(57 \%)} \\
      \footnotesize{Equivalent sentence pairs} & \footnotesize{429} \scriptsize{(43 \%)} \\[1.3mm]
      \footnotesize{Mean sentence word length} & \footnotesize{10.75} \scriptsize{(4.18)} \\
      \footnotesize{Mean span word length} & \footnotesize{3.91} \scriptsize{(2.71)} \\
      \footnotesize{Mean spans per sentence} & \footnotesize{1.30} \scriptsize{(0.51)} \\
      \bottomrule
    \end{tabularx}
    \caption{Global statistics of the SSD. For the word counts, simple whitespace tokenization was used, not considering punctuation.}\label{tab:sdd-statistics}
\end{table}

To ensure the correctness of the semi-automatically annotated data, we ask 6 different annotators with no previous contact with the task to annotate a subset of 100 random samples from the SSD. We then compute all combinations of paired inter-annotator agreements, and calculate their mean and standard deviation. We also report the mean agreement between each of the annotators and the gold samples proceeding from the SSD. The results, included in \autoref{tab:annotation-agreements}, show a substantial level of agreement, indicating that the task is well defined. After performing an error analysis, we detect that the slightly lower agreement on the span counts is likely derived from spans involving conjunctions, such as expressions connected by \q{and} either being included in the same span, or split into two different spans.

\begin{table}
  \centering
  \begin{tabular}{lcc}
    \toprule
    {\fontsize{8}{12}\selectfont \textbf{Agreement}} & {\fontsize{8}{12}\selectfont \textbf{Inter-annotator}} & {\fontsize{8}{12}\selectfont \textbf{Annotator-dataset}} \\
    \midrule
      {\fontsize{7.5}{10}\selectfont Span boundaries} & {\fontsize{7.5}{10}\selectfont 0.74} \scriptsize{(0.05)} & {\fontsize{7.5}{10}\selectfont 0.67} \scriptsize{(0.03)} \\
      {\fontsize{7.5}{10}\selectfont Span labels} & {\fontsize{7.5}{10}\selectfont 0.87} \scriptsize{(0.05)} & {\fontsize{7.5}{10}\selectfont 0.91} \scriptsize{(0.04)} \\
      {\fontsize{7.5}{10}\selectfont Span count} & {\fontsize{7.5}{10}\selectfont 0.66} \scriptsize{(0.06)} & {\fontsize{7.5}{10}\selectfont 0.61} \scriptsize{(0.07)} \\
      \bottomrule
    \end{tabular}
    \caption{Mean Cohen's $\kappa$ scores on different annotation aspects (standard deviations in parenthesis).}\label{tab:annotation-agreements}
\end{table}

\section{Experimental Setup}

Next, we present and explain the different strategies we considered to tackle the problem of DSD. We also introduce the evaluation schema adopted to assess their performance.

\subsection{Methods}

We propose a total of 5 methods plus 2 baselines. Some of these methods have the advantage of working with smaller models and requiring no fine-tuning at all, and most are, to a significant extent, model agnostic. We also consider two dummy baselines that require no model.

\vspace{2mm}

\textbf{SHAP-DSD} \> This approach is powered by the use of SHAP (SHapley Additive exPlanations). SHAP is a popular framework that allows users to obtain explanations of individual model predictions. It is based on game theory principles, building on the concept of Shapley values \citep{shapley1953}, where the central idea, when applied to Machine Learning, is to fairly distribute an importance value among a set of input features that contribute to the model's prediction.

SHAP's \texttt{Explainer} class is the central \q{explainer} within the framework. In our case, instead of initializing it directly with a model, we pass a callable (function) \textit{f} that takes in a pair of strings and returns the \textit{dissimilarity score} between their embeddings, calculated as:

\vspace{-0.3cm}
\begin{equation}
    \label{eq:dissimilarity-score}
    \scalebox{0.93}{%
        \textrm{DissimilarityScore}(a, b) = 1 - \textrm{CosineSimilarity}(a, b)
    }
\end{equation}
\vspace{0.04cm}

There exist a myriad of models we can use to encode the text. We decided to choose models within the Sentence Transformers family \citep{reimers-2019-sentence-bert}, since they are suited to work with sentences, matching the nature of our dataset. Nonetheless, the advantage of basing our explanations solely on the dissimilarity score is that we can plug in \textit{any} embedding model, making this method very flexible and easy to update as new advancements in the field are made.

Given the two sentences, we run the explainer and obtain the SHAP values for each of the tokens in the second sentence. If the SHAP value (i.e., dissimilarity value) for a token is above a certain dissimilarity threshold,\footnote{This threshold is dependent on the underlying model employed.} we consider that token to be dissimilar. Contiguous tokens classified as dissimilar are then included in the same span.

The caveat of this method is that the annotations will only be as good as the model's sensitivity to semantic dissimilarity. This is the case with all the unsupervised methods we propose.

\vspace{2mm}

\textbf{LIME-DSD} \> The core idea behind LIME (\textit{Local Interpretable Model-agnostic Explanations}) is to locally explain a black-box model through a simpler, transparent model that can be interpreted by humans. \textit{Locally} here means that the simpler surrogate model can only provide explanations on the vicinity of the instance being predicted. It achieves this by training the surrogate model on perturbated versions of the instance being explained. In the case of textual data, the perturbations consist in dropping or masking words or characters in the input text.

We proceed in a similar fashion as with SHAP-DSD, using also Sentence Transformer models. Instead of SHAP values, we now work with the explanation weights predicted by LIME for each token. We then enclose within spans those tokens with weights above a threshold.

\vspace{2mm}

\textbf{Embedding-DSD} \> This method constitutes a novel contribution. It bears close similarities to SHAP or LIME in that it aims to explain individual samples by applying certain perturbations on them. However, in our case, we intentionally aim to develop a method that, despite being less generally applicable, works better within the specific context of DSD. For that, instead of relying on masking, or even entirely dropping tokens, our approach considers all possible \textit{n}-gram replacements that can be made from the first sentence to the second sentence, and then calculates the impact each of these replacements has on the overall similarity of the input pair.

Let us illustrate with an example why this might be a more effective perturbation when dealing with DSD. Imagine we are given two sentences that we want to compare:

\begin{itemize}[itemsep=0.1pt,topsep=1pt]
    \item \texttt{the bird flies fast over the hill} \\[-6mm]
    \item \texttt{the car rides fast over the hill}
\end{itemize}

\begin{figure}[ht]
  \includegraphics[width=\columnwidth]{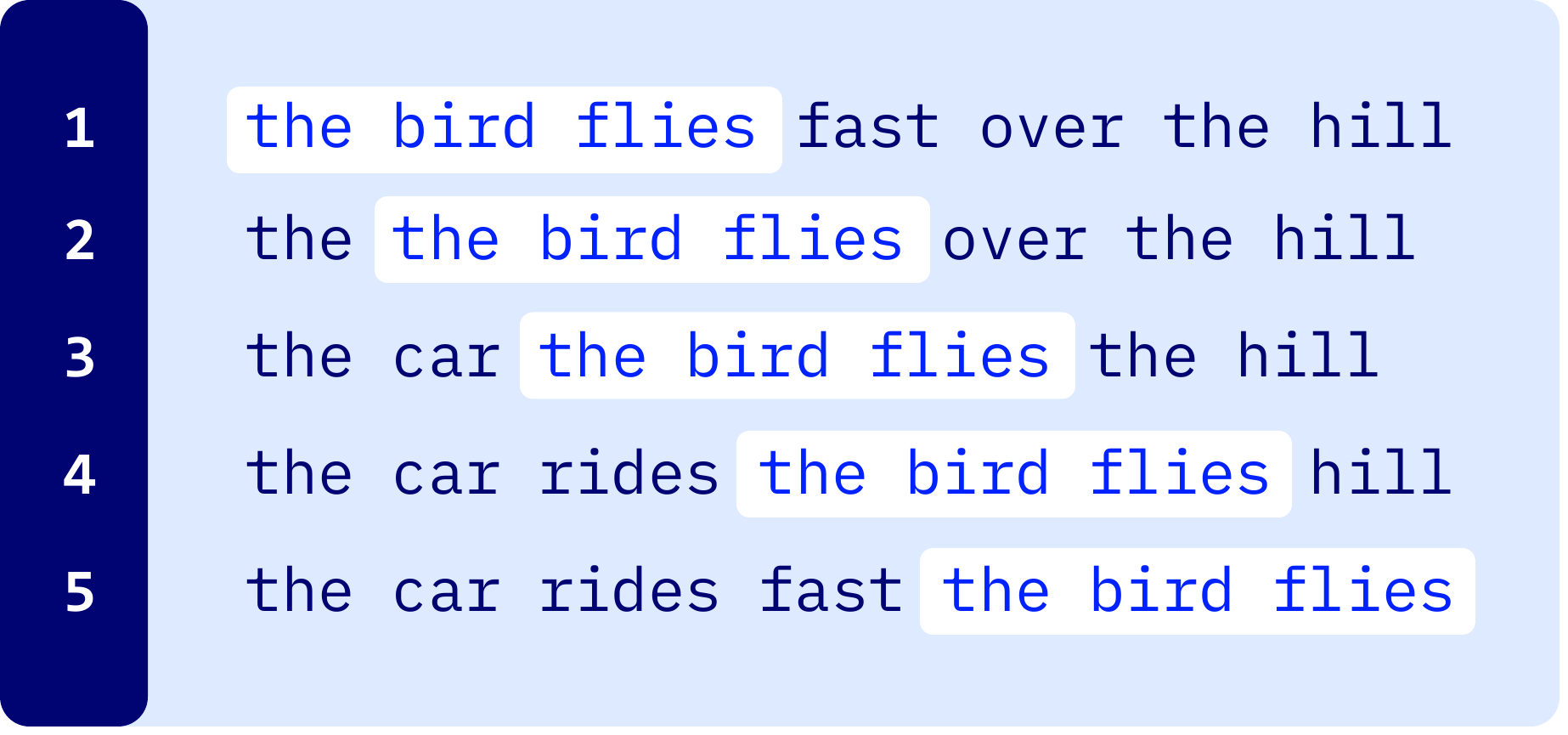}
  \caption{Example of trigram replacements. The considered trigrams proceeding from the first sentence (highlighted) are inserted into the second sentence, replacing the original trigrams.}
  \label{fig:embedding-sentence-diff}
\end{figure}

First, we calculate the \textit{base similarity} between the sentences, which is simply their cosine similarity (or any other similarity measure). In our example, let us say that this base similarity is 0.6. We then compute all the unigrams in the first sentence, followed by all the bigrams, trigrams, etc., until we reach an \textit{n}-gram that spans the entirety of the sentence, i.e., when \textit{n} is equal to the number of unigrams in it. Next, we replace \textit{all} the obtained \textit{n}-grams from the previous step in \textit{all} possible positions in the second sentence. If, for example, we were considering the trigram \texttt{[\textquotesingle the\textquotesingle, \textquotesingle bird\textquotesingle, \textquotesingle flies\textquotesingle]} from the first sentence, we would obtain the replacements in the second sentence depicted in \autoref{fig:embedding-sentence-diff}.

We then calculate the similarity between each of the replacements and the first, original sentence, and identify which one of the replacements achieves the highest similarity gain compared to the initial base similarity. Coming back to our example, we can see that the first replacement is identical to the first original sentence, resulting in a similarity of 1.0. Its similarity gain is then 0.4, calculated as 1.0 - 0.6 (base similarity). We then associate this value with each unigram in the trigram (i.e., \texttt{\textquotesingle the\textquotesingle}, \texttt{\textquotesingle bird\textquotesingle}, \texttt{\textquotesingle flies\textquotesingle}). By the end of the iterations, each unigram in the second sentence will have a series of similarity gains from all \textit{n}-gram replacements that include it.

The similarity gains associated to each of the unigrams are combined through an \textit{aggregation function}. In our case, we define a function that assigns smaller weights to gains coming from higher \textit{n}-grams, in order to pay more attention to local differences:

\vspace*{-0.25cm}
\begin{equation}
    \label{eq:aggregated-gain}
    AggGain_{unigram} = \frac{1}{n} \cdot \sum_{i=1}^{n} \frac{gains_i}{i} \>,
\end{equation}

\noindent where $gains$ is an ordered array containing the gains coming from the unigrams, bigrams, \ldots, \textit{n}-grams in which the unigram appears, and $n$ is the length of this array. This function will assign less weight to gains corresponding to higher \textit{n}-grams (i.e., those that appear later in the $gains$ array). We normalize the aggregated gains by $n$, since unigrams at the beginning and end of the sentence will appear in less \textit{n}-grams, and therefore will have a smaller amount of gain values associated to them.

Once we have a final, aggregated gain for each unigram, we apply a threshold and those unigrams with gains above it are considered dissimilar, grouping contiguous, dissimilar unigrams in the same span. \autoref{sec:appendix-embedding-dsd-algorithm} includes a pseudocode version of our algorithm.

This method, once again, relies on a good embedding model to work optimally. In this case, we employed not only Sentence Transformer models, but also commercially available embedding models.\footnote{Models from OpenAI and Google were used.} This allowed us to draw a comparison between small (former) and large embedding models (latter), and determine whether the performance gain, or lack thereof, is worth the extra cost.

\vspace{2mm}

\textbf{LLM-DSD} \> Recently, the NLP field has experienced notable advancements due to the rapid development of Large Language Models (LLMs). These models have shown outstanding performance in zero-shot or few-shot tasks, so they constituted an evident approach for our task.

When working with LLMs, a carefully designed prompt is crucial to obtain good results, especially when working in zero-shot or few-shot scenarios \citep{zhao2023survey}, which is also our case. We formulate our prompt as a 4-shot learning task, providing 2 examples of correct annotations, and 2 examples of incorrect annotations. We employed models from a variety of vendors, namely OpenAI, Meta, Anthropic and DeepSeek.

\vspace{2mm}

\textbf{Token-Classification-DSD} \> This method approaches the problem as a token classification task (also known as sequence labeling), similarly to how Part-of-speech tagging or Named Entity Recognition (NER) are commonly dealt with. In order to transform the data into a suitable format for token classification, we tokenize the training input samples and annotate them following the BIO tagging schema \citep{ramshaw-marcus-1995-text}: the first token within a span is tagged with a \q{B}, the rest of the tokens within the span are tagged with an \q{I}, and any other token (i.e., tokens outside any span) are labeled with an \q{O}. It is important to note that we only attend to spans that are dissimilar, that is, those labeled as 0 in the SSD. Annotated span markers for  equivalent spans are removed, and we do not label them in any other way.

Instead of training from scratch, we fine-tuned BERT and RoBERTa \citep{Liu2019RoBERTaAR} models, as well as their distilled counterparts, DistilBERT and DistilRoBERTa \citep{sanh2020distilbert}, on the SSD. We use cross-entropy loss, with a learning rate $\eta = \text{5} \cdot \text{10}^\text{-5}$, a weight decay of $\text{5} \cdot \text{10}^\text{-3}$, and a batch size of 8. We adopt a 5-fold cross-validation setting, training the models for 5 epochs on 4 of the folds and evaluating the fine-tuned model on the remaining fold.

\vspace{2mm}

\textbf{Baseline Methods} \> In addition to the previous methods, we propose two simple baselines, No-DSD and Naive-DSD, exploiting certain peculiarities of the SSD. These two techniques have no notion of meaning and are very fast to compute.

\textbf{\small{No-DSD}} \, Here, we just return the second sentence as it is, with no dissimilar spans annotated. This is presumably a strong baseline due to the fact that roughly half of the sentence pairs in the SSD contain no dissimilar spans. Therefore, this method will always annotate such pairs accurately. On the other hand, no pair actually containing dissimilar spans will get correctly annotated.

\textbf{\small{Naive-DSD}} \, In this case, any word in the second sentence that does not appear in the first sentence is considered as dissimilar. This method will fail for those equivalent spans that convey the same meaning but employ a different wording. However, annotations will be fairly accurate for spans that are indeed dissimilar.\footnote{Note that dissimilar span pairs might still contain common words, such as articles or prepositions.}

\begin{table*}
  \centering
  \footnotesize
  \setlength\tabcolsep{0pt}
  \setlength\extrarowheight{6pt}
  \begin{tabular*}{\textwidth}{@{}m{3.5cm}@{\extracolsep{\fill}}*{7}{c}@{}}
    \toprule
      & & \multicolumn{4}{c}{\footnotesize{\textbf{SSD}}} & \multicolumn{2}{c}{\footnotesize{\textbf{SemEval-2016}}} \\ \cline{3-6} \cline{7-8}

      \textbf{Method} & \textbf{Model Size} & \textbf{F1-Global} & \textbf{F1-NoDiff} & \textbf{F1-Diff} & \textbf{Eval. Time} & \textbf{F1-Diff} & \textbf{Eval. Time} \\
    \midrule
      LIME \vspace{0.15cm}\newline \small{\texttt{all-mpnet-base-v2}} \, \tiny{(0.001)} & 109M & 0.463 & 0.782 & 0.223 & 1981.81 & 0.109 & 199.89 \\

      SHAP \vspace{0.15cm}\newline \small{\texttt{all-MiniLM-L6-v2}} \, \tiny{(0.010)} & 22.7M & 0.366 & 0.434 & 0.131 & 44.47 & 0.306 & 4.52 \\

      Embedding \vspace{0.15cm}\newline \small{\texttt{all-mpnet-base-v2}} \, \tiny{(0.006)} & 109M & 0.469 & 0.529 & 0.423 & 11.28 & 0.352 & 1.13 \\

      Embedding \vspace{0.15cm}\newline \small{\texttt{text-embedding-004}} \tiny{(0.005)} & Billions & 0.547 & 0.666 & 0.459 & 611.42 & 0.338 & 34.57 \\
    
      LLM \newline \small{\texttt{Claude 3.5 Sonnet}} & Billions & \textbf{0.750} & \textbf{0.895} & \textbf{0.640} & 337.80 & 0.389 & 39.22 \\

      Token Classification \newline \small{\texttt{roberta-base}} & 125M & 0.690 & 0.838 & 0.574 & \textbf{1.23} & \textbf{0.441} & \textbf{0.10} \\
    \midrule
      No-DSD & N/A & 0.429 & 1.000 & 0.000 & N/A & 0.000 & N/A \\

      Naive-DSD & N/A & 0.311 & 0.003 & 0.542 & N/A & 0.307 & N/A \\
    \bottomrule
  \end{tabular*}
  \caption{Summary of the results. LLMs and supervised methods (Token-Classification-DSD) perform best, followed by our proposed method (Embedding-DSD). For methods that require a threshold, we include it next to the model's name. Evaluation times are reported in minutes. For experiments relying on external APIs (\texttt{text-embedding-004} and Claude 3.5 Sonnet), times might vary depending on rate limits, connection speeds, etc. For these models, the exact number of parameters remains undisclosed, so we are unable to report it. The rest of experiments were run on a NVIDIA RTX 3080Ti graphics card.}\label{tab:methods-comparison-table}
\end{table*}

\subsection{Evaluation}

We evaluate the different methods both on our dataset, as well as the subset of samples from the SemEval-2016 Task 2 dataset with spans annotated as \textit{opposite}. For a more robust evaluation given the relatively small size of the datasets, we adopt a 5-fold cross-validation scheme. For the unsupervised methods, we simply evaluate on the test fold, leaving the remaining 4 folds unused.

Considering that the number of annotated spans can be different from the number of reference spans, we first perform \textit{span alignment}. For that, given that span pairs in the SSD are ordered, we align the reference and annotated spans in a way that the offset between them is minimized. This offset is calculated as follows:

\vspace{-0.62cm}
\begin{multline}
    \label{eq:span-offset}
        {\fontsize{10}{12}\selectfont \text{Offset}(s_{ref}, s_{ann})} = \\ {\fontsize{7}{10}\selectfont | s_{ref}.\text{start} - s_{ann}.\text{start}  | \> + \> | s_{ref}.\text{end} - s_{ann}.\text{end} |} \> ,
\end{multline}

\noindent where $s_{ref}$ and $s_{ann}$ are the reference and annotated spans, respectively, and \textit{start} and \textit{end} denote the starting and ending indexes of the span. Since the number of spans per sentence is low, we compute all the possible alignments and choose the one that results in the smallest offset.

For each of the aligned spans, we calculate, at the unigram level, precision, recall, and F1-score in a similar fashion to the METEOR metric \citep{denkowski:lavie:meteor-wmt:2014}. Scores are averaged across all spans in the input pair, giving result to the overall precision, recall and F1-score for that sample. Given that we only get annotations for the second sentence, we then swap them to also evaluate the annotation of the first sentence.

\subsubsection{Downstream Task}

Additionally, we set up an experiment employing DSD for the task of paraphrase detection. This serves as a plausibility check to ensure the correctness and suitability of DSD in a real-world scenario. The evaluation is performed on the test split of the PAWS-Wiki Labeled (Final) Corpus \citeplanguageresource{paws2019naacl}, containing pairs of sentences labeled with either 0 (no paraphrase) or 1 (paraphrase).

First, we use a Sentence Transformer model to perform STS. In order to adapt the model to the task, we finetune it on the training split of the PAWS-Wiki Labeled dataset. At inference time, we label a pair as a paraphrase if the similarity is equal or greater than a specific threshold. We find the optimal threshold for each model by evaluating on the validation set of the dataset. Once found, we run a final evaluation on the test split.

Then, we repeat the experiment, but only label a pair as paraphrase if the pre-established similarity threshold is reached \textit{and} the pair contains no dissimilar spans. These spans are obtained by applying Embedding-DSD on the same model used for STS.

In both setups we calculate the accuracy as the number of correctly classified pairs divided by the total number of pairs. The goal is to determine whether introducing DSD leads to increased accuracies in paraphrase identification.

\section{Results and Discussion}\label{sec:results-and-discussion}

In the case of the SSD, in order to account for false positives (i.e., spans labeled as dissimilar when they are not), we report separate metrics for those sentence-pairs that contain no dissimilar spans (\textit{NoDiff}), and those that do (\textit{Diff}). This distinction aims to determine how often evaluated methods treat semantically equivalent spans with lexical differences as dissimilar. We also report results taking into consideration both the \textit{Diff} and \textit{NoDiff} scores (\textit{Global}). As for the SemEval-2016 data, we only report the \textit{Diff} scores, since we only consider samples with dissimilar spans.

\autoref{tab:methods-comparison-table} collects the best results from each of the presented methods. LLMs and supervised methods (i.e., token classification) obtain the highest scores overall. However, the latter is much faster on average, employing models orders of magnitude smaller (ranging around 100 million parameters), at the cost of requiring to be fine-tuned on the task. Our proposed method, Embedding-DSD, achieves better results than the other two considered explainability methods, based on LIME and SHAP, while having considerably lower inference times.

The results for the SemEval-2016 data must be taken as a lower bound, since we only considered spans labeled as \textit{opposite}. As mentioned earlier, other types of spans still remain that are in fact dissimilar, leading to lower scores when compared to the STS results.

Overall, these results prove the difficulty of the task. Specifically, methods that rely on embedding models show the general lack of sensitivity of such models towards localized textual changes. In \autoref{sec:appendix-annotated-output-examples}, we include examples of annotations produced by these and the other methods. The annotation schema introduced in this work and used for the SSD could be used to fine-tune STS models that are more sensitive to small lexical variations. This is something that has been proven to work in the case of negations \citep{anschütz2023correct}.

\autoref{tab:results-paws-wiki} includes the results from two different models when applying DSD to the concrete task of paraphrase detection. Here, the introduction of DSD improves accuracy up to 8 points. It is worth noting that this improvement is achieved without any modifications to the base model or further finetuning on the task.

To conclude this discussion, we would like to mention that, although not the main focus of this work, DSD could also be used to evaluate models' sensitivity to dissimilarity. This evaluation could serve as a complement to STS metrics, or be used as an objective when training or finetuning models for STS related tasks, potentially leading to models that capture semantics in a more complete way.

\section{Conclusion}

In this work, we present the task of Dissimilar Span Detection (DSD): a method to improve the interpretability and reliability of STS scores. The task consists in, given two texts, identifying spans pairs with a common semantic function, but conveying different meanings. DSD can complement current STS metrics, which typically report a single score, providing interpretability in user-facing scenarios. Additionally, it can potentially enhance the performance of current approaches that rely on STS.

To lay the foundation for the task, we introduce an appropriate annotation and evaluation schema and release the Span Similarity Dataset (SSD). This dataset, developed via a semi-automated pipeline that integrates LLMs with human verification, serves as a valuable starting point for the research community, facilitating further advancements in the field.

We propose and evaluate a number of methods, both unsupervised and model-agnostic, based on LIME, SHAP, LLMs, and our own method, as well as a further supervised approach. These approaches are conceived as a first attempt to tackle the task of DSD, and should therefore be taken as baselines. Regardless, the obtained results reflect the challenging nature of the task. However, even in its current state, DSD can already increase performance in the specific task of paraphrase detection. Nevertheless, there is still ample room for refinement in future work. Immediate improvements can be applied to a number of different fronts: the SSD dataset can be expanded with further samples; approaches that build on the presented methods, or other new approaches, can be explored; finally, DSD can be further integrated and tested on different downstream tasks apart from paraphrase detection.

\begin{table}
    \centering
    \begin{tabularx}{\columnwidth}{Xcc}
        \toprule
        {\textbf{Model}} & \scriptsize{\textbf{STS}} & \scriptsize{\textbf{STS+DSD}} \\
        \midrule
        \small{\texttt{all-MiniLM-L6-v2}} \, \tiny{(0.65, 8$\cdot$10\textsuperscript{-3})} & \small{0.720} & \textbf{\small{0.808}} \\
        \small{\texttt{all-mpnet-base-v2}} \, \tiny{(0.63, 8$\cdot$10\textsuperscript{-3})} & \small{0.795} & \textbf{\small{0.868}} \\
        \bottomrule
    \end{tabularx}
\caption{Comparison of accuracies on the PAWS-Wiki Labeled using uniquely STS or combining STS with DSD. We include the STS and DSD thresholds in parenthesis next to the models.}\label{tab:results-paws-wiki}
\end{table}

\section{Limitations}

The main limitations of our work stem from aspects regarding the annotation of the SSD, namely:

\begin{itemize}[itemsep=0.1pt,topsep=1pt]
    \item It currently contains data solely in English. In future work, a multilingual setting could be considered. Still, all the methods presented here would be applicable with no or minor modifications.
    \item We worked exclusively at the sentence level. An even more challenging setup would work with full paragraphs, or even longer textual units. A baseline approach would consist in 1) splitting the text into sentences, 2) performing sentence alignment, so that the most similar sentences are paired together, and 3) running any of our developed methods.
    \item In the dataset, corresponding spans always appear in the same order in both the first and the second sentence. In future work, a more complete setup would allow span pairs to come in any order.
    \item Our methods focused solely on DSD, but a full-fledged system would also be able to identify equivalent spans. Furthermore, it could also be possible to include spans that only appear in one of the sentences. This setting would enable support for further tasks, such as hallucination detection.
\end{itemize}

These limitations are a byproduct of the approach we adopted towards the task of DSD: establish a common framework, but start with a more restricted and simplified setting in order to lay solid foundations that can be expanded in future work.

\section*{Ethical Considerations}

The main goal of our work is to provide techniques and foster new ideas that help improve the interpretability and reliability of present STS metrics. We are currently unable to envision ways in which our work can be maliciously misused, although we welcome any suggestion and commit to continuously reviewing our methodologies in order to ensure a safe, future development of the task.

We are aware, however, that the presented methods still perform at an unacceptable level for any kind of real-world deployment, and strongly advocate against them being used in production until further research has been conducted and performance and reliability have been improved.

Finally, since the SSD dataset has been crafted from an amalgam of previously released datasets, it may contain biases already present in them. If any of these biases became evident at any point in the future, we remain fully committed to reviewing and rectifying our dataset.

\section*{Acknowledgments}

The work was funded/co-funded by the European Union (ERC, EPICAL, 101141712). Views and opinions expressed are however those of the author(s) only and do not necessarily reflect those of the European Union or the European Research Council. Neither the European Union nor the granting authority can be held responsible for them.

Besides, we would like to thank the six annotators that provided their invaluable time to contribute to this work.

\section{Bibliographical References}\label{sec:reference}

\bibliographystyle{lrec2026-natbib}
\bibliography{lrec2026}

\section{Language Resource References}\label{lr:ref}
\bibliographystylelanguageresource{lrec2026-natbib}
\bibliographylanguageresource{languageresource}

\onecolumn
\appendix

\section{Detailed Results}
\label{sec:appendix-detailed-results}

\autoref{tab:semeval-2016-results} and \autoref{tab:ssd-results} include a detailed collection of the results obtained by each of the different proposed DSD methods.

For the SSD, we report precision, recall and F1 scores evaluated on both sentence-pairs containing no differing spans (i.e., \textit{NoDiff}), as well as on pairs including such spans (i.e., \textit{Diff}). We also report the global performance considering both scenarios (i.e., \textit{Global}).

For the SemEval-2016 Task 2 data, we only report the \textit{NoDiff} scores, since we only consider sentences that contain \textit{opposite} spans.\\[1cm]

\begin{table}[H]
  \centering
  \setlength\tabcolsep{0pt}
  \setlength\extrarowheight{2pt}
  \begin{tabular*}{\textwidth}{@{\extracolsep{\fill}}*{2}{l}@{\extracolsep{\fill}}*{4}{c}}
    \toprule
      & & \multicolumn{3}{c}{\footnotesize{\textbf{NoDiff}}} & \multirow{2}*{\textbf{\footnotesize Time}} \\ \cline{3-5}

      \footnotesize{\textbf{Method}} & \footnotesize{\textbf{Model}} & \footnotesize{\textbf{Precision}} & \footnotesize{\textbf{Recall}} & \footnotesize{\textbf{F1}} & \scriptsize{(minutes)} \\
    \midrule
      \multirow{2}*{\footnotesize{LIME}} & \small{\texttt{all-MiniLM-L6-v2}} \hspace{1.5mm} \scriptsize{(0.010)} & \footnotesize{\textbf{0.253}} \scriptsize{(0.010)} & \footnotesize{\textbf{0.395}} \scriptsize{(0.029)} & \footnotesize{\textbf{0.269}} \scriptsize{(0.013)} & \footnotesize{\textbf{120.88}} \\

      & \small{\texttt{all-mpnet-base-v2}} \hspace{1.5mm} \scriptsize{(0.001)} & \footnotesize{0.154} \scriptsize{(0.017)} & \footnotesize{0.093} \scriptsize{(0.011)} & \footnotesize{0.109} \scriptsize{(0.011)} & \footnotesize{199.89} \\
    \midrule
      \multirow{2}*{\footnotesize{SHAP}} & \small{\texttt{all-MiniLM-L6-v2}} \hspace{1.5mm} \scriptsize{(0.005)} & \footnotesize{\textbf{0.247}} \scriptsize{(0.012)} & \footnotesize{\textbf{0.529}} \scriptsize{(0.025)} & \footnotesize{\textbf{0.306}} \scriptsize{(0.016)} & \footnotesize{\textbf{4.52}} \\

      & \small{\texttt{all-mpnet-base-v2}} \hspace{1.5mm} \scriptsize{(0.001)} & \footnotesize{0.243} \scriptsize{(0.016)} & \footnotesize{0.480} \scriptsize{(0.022)} & \footnotesize{0.293} \scriptsize{(0.019)} & \footnotesize{21.58} \\
    \midrule
      \multirow{4}*{\footnotesize{Embedding}} & \small{\texttt{all-MiniLM-L6-v2}} \hspace{1.5mm} \scriptsize{(0.005)} & \footnotesize{\textbf{0.249}} \scriptsize{(0.015)} & \footnotesize{\underline{\textbf{0.842}}} \scriptsize{(0.051)} & \footnotesize{\textbf{0.362}} \scriptsize{(0.019)} & \footnotesize{\textbf{0.38}} \\

      & \small{\texttt{all-mpnet-base-v2}} \hspace{1.5mm} \scriptsize{(0.005)} & \footnotesize{0.243} \scriptsize{(0.013)} & \footnotesize{0.823} \scriptsize{(0.064)} & \footnotesize{0.352} \scriptsize{(0.017)} & \footnotesize{1.13} \\[1.5mm]

      & \small{\texttt{text-embedding-3-large}} \hspace{1.5mm} \scriptsize{(0.005)} & \footnotesize{0.248} \scriptsize{(0.015)} & \footnotesize{0.829} \scriptsize{(0.056)} & \footnotesize{0.355} \scriptsize{(0.020)} & \footnotesize{11.16} \\

      & \small{\texttt{text-embedding-004}} \hspace{1.5mm} \scriptsize{(0.005)} & \footnotesize{0.239} \scriptsize{(0.012)} & \footnotesize{0.778} \scriptsize{(0.007)} & \footnotesize{0.338} \scriptsize{(0.018)} & \footnotesize{34.57} \\
    \midrule
      \multirow{4}*{\footnotesize{LLM}} & \footnotesize{GPT-4o \hspace{1mm} \texttt{2024-11-20}} & \footnotesize{0.376} \scriptsize{(0.031)} & \footnotesize{\textbf{0.609}} \scriptsize{(0.033)} & \footnotesize{\textbf{0.416}} \scriptsize{(0.031)} & \footnotesize{45.22} \\

      & \footnotesize{Llama 3.3 70B Instruct \hspace{1mm} \texttt{2024-12-06}} & \footnotesize{\underline{\textbf{0.409}}} \scriptsize{(0.045)} & \footnotesize{0.530} \scriptsize{(0.054)} & \footnotesize{0.413} \scriptsize{(0.047)} & \footnotesize{92.86} \\

      & \footnotesize{Claude 3.5 Sonnet \hspace{1mm} \texttt{2024-10-22}} & \footnotesize{0.344} \scriptsize{(0.040)} & \footnotesize{0.593} \scriptsize{(0.041)} & \footnotesize{0.389} \scriptsize{(0.040)} & \footnotesize{\textbf{39.22}} \\

      & \footnotesize{DeepSeek R1 \hspace{1mm} \texttt{2025-01-20}} & \footnotesize{0.379} \scriptsize{(0.048)} & \footnotesize{0.471} \scriptsize{(0.049)} & \footnotesize{0.390} \scriptsize{(0.050)} & \footnotesize{1585.06} \\
    \midrule
      \multirow{4}*{\footnotesize{\shortstack[l]{Token \\ Classification}}} & \small{\texttt{distilbert-base-uncased}} & \footnotesize{0.379} \scriptsize{(0.063)} & \footnotesize{0.517} \scriptsize{(0.083)} & \footnotesize{0.424} \scriptsize{(0.069)} & \footnotesize{\underline{\textbf{0.06}}} \\

      & \small{\texttt{bert-base-uncased}} & \footnotesize{\textbf{0.390}} \scriptsize{(0.023)} & \footnotesize{0.559} \scriptsize{(0.019)}  & \footnotesize{\underline{\textbf{0.446}}} \scriptsize{(0.022)}  & \footnotesize{0.11} \\[1.5mm]

      & \small{\texttt{distilroberta-base}} & \footnotesize{0.374} \scriptsize{(0.033)} & \footnotesize{\textbf{0.586}} \scriptsize{(0.024)} & \footnotesize{0.444} \scriptsize{(0.029)} & \footnotesize{0.62} \\

      & \small{\texttt{roberta-base}} & \footnotesize{0.369} \scriptsize{(0.022)} & \footnotesize{0.579} \scriptsize{(0.018)} & \footnotesize{0.441} \scriptsize{(0.020)} & \footnotesize{0.10} \\
    \midrule
      \multirow{2}*{\footnotesize{Baselines}} & \footnotesize{No-DSD} & \footnotesize{0.000} \scriptsize{(0.000)} & \footnotesize{0.000} \scriptsize{(0.000)} & \footnotesize{0.000} \scriptsize{(0.000)} & \footnotesize{N/A} \\

      & \footnotesize{Naive-DSD} & \footnotesize{\textbf{0.317}} \scriptsize{(0.023)} & \footnotesize{\textbf{0.389}} \scriptsize{(0.039)} & \footnotesize{\textbf{0.307}} \scriptsize{(0.028)} & \footnotesize{N/A} \\
    \bottomrule
  \end{tabular*}
  \caption{Results of the different methods on the SemEval-2016 Task 2 data. As previously mentioned, we only consider the sentences that contain spans annotated as \textit{opposite}. That is why we only report the \textit{NoDiff} scores. We include standard errors within parenthesis. For methods that require a threshold, its value is included in parenthesis next to the model's name. Best results from each method are shown in bold, while best results overall in a column are also underlined. The time column reflects the total time to run the entire evaluation (i.e., 5 folds). Note that for methods that rely on external APIs (e.g., LLM-DSD), times might fluctuate depending on rate limits, connection speeds, etc.}\label{tab:semeval-2016-results}
\end{table}

\begin{sidewaystable}
  \centering
  \begin{tabularx}{0.95\textwidth}{lXccc|ccc|ccc|c}
    \toprule
      \multirow{2}*{\textbf{\footnotesize Method}} & \multirow{2}*{\textbf{\footnotesize Model}} & \multicolumn{3}{c|}{\textbf{\footnotesize Precision}} & \multicolumn{3}{c|}{\textbf{\footnotesize Recall}} & \multicolumn{3}{c|}{\textbf{F1}} & \multirow{2}*{\begin{tabular}{@{}c@{}} \textbf{\footnotesize Time} \vspace{-1.8mm} \\ \scriptsize{(minutes)} \end{tabular}} \\ & & \textbf{\fontsize{8}{10}\selectfont Global} & \textbf{\fontsize{8}{10}\selectfont NoDiff} & \textbf{\fontsize{8}{10}\selectfont Diff} & \textbf{\fontsize{8}{10}\selectfont Global} & \textbf{\fontsize{8}{10}\selectfont NoDiff} & \textbf{\fontsize{8}{10}\selectfont Diff} & \textbf{\fontsize{8}{10}\selectfont Global} & \textbf{\fontsize{8}{10}\selectfont NoDiff} & \textbf{\fontsize{8}{10}\selectfont Diff} & \\
    \midrule
      \footnotesize{LIME} & \begin{tabular}{@{}l@{}} \small{\texttt{all-MiniLM-L6-v2}} \, \tiny{(0.030)} \\ \small{\texttt{all-mpnet-base-v2}} \, \tiny{(0.001)} \end{tabular} & \begin{tabular}{@{}c@{}} \scriptsize{0.355} \tiny{(0.003)} \\ \scriptsize{\textbf{0.563}} \tiny{(0.009)} \end{tabular} & \begin{tabular}{@{}c@{}} \scriptsize{0.371} \tiny{(0.012)} \\ \scriptsize{\textbf{0.782}} \tiny{(0.014)} \end{tabular} & \begin{tabular}{@{}c@{}} \scriptsize{0.342} \tiny{(0.011)} \\ \scriptsize{\textbf{0.397}} \tiny{(0.016)} \end{tabular} & \begin{tabular}{@{}c@{}} \scriptsize{0.287} \tiny{(0.005)} \\ \scriptsize{\textbf{0.436}} \tiny{(0.014)} \end{tabular} & \begin{tabular}{@{}c@{}} \scriptsize{0.371} \tiny{(0.012)} \\ \scriptsize{\textbf{0.782}} \tiny{(0.014)} \end{tabular} & \begin{tabular}{@{}c@{}} \scriptsize{\textbf{0.224}} \tiny{(0.009)} \\ \scriptsize{0.177} \tiny{(0.013)} \end{tabular} & \begin{tabular}{@{}c@{}} \scriptsize{0.294} \tiny{(0.004)} \\ \scriptsize{\textbf{0.463}} \tiny{(0.013)} \end{tabular} & \begin{tabular}{@{}c@{}} \scriptsize{0.371} \tiny{(0.012)} \\ \scriptsize{\textbf{0.782}} \tiny{(0.014)} \end{tabular} & \begin{tabular}{@{}c@{}} \scriptsize{\textbf{0.236}} \tiny{(0.010)} \\ \scriptsize{0.223} \tiny{(0.014)} \end{tabular} & \begin{tabular}{@{}c@{}} \scriptsize{\textbf{1018.70}} \\ \scriptsize{1981.81} \end{tabular} \\
    \midrule
      \footnotesize{SHAP} & \begin{tabular}{@{}l@{}} \small{\texttt{all-MiniLM-L6-v2}} \, \tiny{(0.010)} \\ \small{\texttt{all-mpnet-base-v2}} \, \tiny{(0.030)} \end{tabular} & \begin{tabular}{@{}c@{}} \scriptsize{\textbf{0.394}} \tiny{(0.003)} \\ \scriptsize{0.321} \tiny{(0.003)} \end{tabular} & \begin{tabular}{@{}c@{}} \scriptsize{\textbf{0.434}} \tiny{(0.011)} \\ \scriptsize{0.356} \tiny{(0.011)} \end{tabular} & \begin{tabular}{@{}c@{}} \scriptsize{\textbf{0.362}} \tiny{(0.011)} \\ \scriptsize{0.293} \tiny{(0.008)} \end{tabular} & \begin{tabular}{@{}c@{}} \scriptsize{\textbf{0.370}} \tiny{(0.004)} \\ \scriptsize{0.256} \tiny{(0.007)} \end{tabular} & \begin{tabular}{@{}c@{}} \scriptsize{\textbf{0.434}} \tiny{(0.011)} \\ \scriptsize{0.356} \tiny{(0.011)} \end{tabular} & \begin{tabular}{@{}c@{}} \scriptsize{\textbf{0.320}} \tiny{(0.011)} \\ \scriptsize{0.185} \tiny{(0.010)} \end{tabular} & \begin{tabular}{@{}c@{}} \scriptsize{\textbf{0.366}} \tiny{(0.002)} \\ \scriptsize{0.266} \tiny{(0.005)} \end{tabular} & \begin{tabular}{@{}c@{}} \scriptsize{\textbf{0.434}} \tiny{(0.011)} \\ \scriptsize{0.356} \tiny{(0.356)} \end{tabular} & \begin{tabular}{@{}c@{}} \scriptsize{\textbf{0.313}} \tiny{(0.010)} \\ \scriptsize{0.198} \tiny{(0.008)} \end{tabular} & \begin{tabular}{@{}c@{}} \scriptsize{\textbf{44.47}} \\ \scriptsize{216.00} \end{tabular} \\
    \midrule
      \footnotesize{Embedding} & \begin{tabular}{@{}l@{}} \small{\texttt{all-MiniLM-L6-v2}} \, \tiny{( 0.010)} \\ \small{\texttt{all-mpnet-base-v2}} \, \tiny{(0.006)} \\[1.7mm] \small{\texttt{text-embedding-3-large}} \, \tiny{(0.005)} \\ \small{\texttt{text-embedding-004}} \, \tiny{(0.005)} \end{tabular} & \begin{tabular}{@{}c@{}} \scriptsize{0.434} \tiny{(0.008)} \\ \scriptsize{0.445} \tiny{(0.008)} \\[1.7mm] \scriptsize{0.454} \tiny{(0.025)} \\ \scriptsize{\textbf{0.527}} \tiny{(0.008)} \end{tabular}  & \begin{tabular}{@{}c@{}} \scriptsize{0.564} \tiny{(0.014)} \\ \scriptsize{0.529} \tiny{(0.019)} \\[1.7mm] \scriptsize{\textbf{0.699}} \tiny{(0.040)} \\ \scriptsize{0.666} \tiny{(0.013)} \end{tabular} & \begin{tabular}{@{}c@{}} \scriptsize{0.333} \tiny{(0.015)} \\ \scriptsize{0.380} \tiny{(0.012)} \\[1.7mm] \scriptsize{0.334} \tiny{(0.033)} \\ \scriptsize{\textbf{0.421}} \tiny{(0.007)} \end{tabular} & \begin{tabular}{@{}c@{}} \scriptsize{0.506} \tiny{(0.008)} \\ \scriptsize{0.585} \tiny{(0.006)} \\[1.7mm] \scriptsize{0.632} \tiny{(0.023)} \\ \scriptsize{\textbf{0.650}} \tiny{(0.005)} \end{tabular} & \begin{tabular}{@{}c@{}} \scriptsize{0.564} \tiny{(0.014)} \\ \scriptsize{0.529} \tiny{(0.019)} \\[1.7mm] \scriptsize{\textbf{0.699}} \tiny{(0.040)} \\ \scriptsize{0.666} \tiny{(0.013)} \end{tabular} & \begin{tabular}{@{}c@{}} \scriptsize{0.460} \tiny{(0.015)} \\ \scriptsize{0.625} \tiny{(0.015)} \\[1.7mm] \scriptsize{0.599} \tiny{(0.027)} \\ \scriptsize{\underline{\textbf{0.635}}} \tiny{(0.011)} \end{tabular} & \begin{tabular}{@{}c@{}} \scriptsize{0.438} \tiny{(0.009)} \\ \scriptsize{0.469} \tiny{(0.005)} \\[1.7mm] \scriptsize{0.486} \tiny{(0.028)} \\ \scriptsize{\textbf{0.547}} \tiny{(0.008)} \end{tabular} & \begin{tabular}{@{}c@{}} \scriptsize{0.564} \tiny{(0.014)} \\ \scriptsize{0.529} \tiny{(0.019)} \\[1.7mm] \scriptsize{\textbf{0.699}} \tiny{(0.040)} \\ \scriptsize{0.666} \tiny{(0.013)} \end{tabular} & \begin{tabular}{@{}c@{}} \scriptsize{0.341} \tiny{(0.013)} \\ \scriptsize{0.423} \tiny{(0.009)} \\[1.7mm] \scriptsize{0.381} \tiny{(0.035)} \\ \scriptsize{\textbf{0.456}} \tiny{(0.006)} \end{tabular} & \begin{tabular}{@{}c@{}} \scriptsize{\textbf{3.75}} \\ \scriptsize{11.28} \\[1.7mm] \scriptsize{111.52} \\ \scriptsize{611.42} \end{tabular} \\
    \midrule
      \footnotesize{LLM} & \begin{tabular}{@{}l@{}} {\fontsize{7}{10}\selectfont GPT-4o \hspace{1mm} \texttt{2024-11-20}} \\ {\fontsize{7}{10}\selectfont Llama 3.3 70B Instruct \hspace{1mm} \texttt{2024-12-06}}  \\ {\fontsize{7}{10}\selectfont Claude 3.5 Sonnet \hspace{1mm} \texttt{2024-10-22}} \\ {\fontsize{7}{10}\selectfont DeepSeek R1 \hspace{1mm} \texttt{2025-01-20}} \end{tabular} & \begin{tabular}{@{}c@{}} \scriptsize{0.854} \tiny{(0.009)} \\ \scriptsize{0.841} \tiny{(0.008)} \\ \scriptsize{\underline{\textbf{0.886}}} \tiny{(0.005)} \\ \scriptsize{0.622} \tiny{(0.011)} \end{tabular} & \begin{tabular}{@{}c@{}} \scriptsize{0.833} \tiny{(0.010)} \\ \scriptsize{0.831} \tiny{(0.012)} \\ \scriptsize{\textbf{0.895}} \tiny{(0.009)} \\ \scriptsize{0.302} \tiny{(0.014)} \end{tabular} & \begin{tabular}{@{}c@{}} \scriptsize{0.870} \tiny{(0.013)} \\ \scriptsize{0.847} \tiny{(0.013)} \\ \scriptsize{\underline{\textbf{0.878}}} \tiny{(0.007)} \\ \scriptsize{0.861} \tiny{(0.007)} \end{tabular} & \begin{tabular}{@{}c@{}} \scriptsize{0.709} \tiny{(0.004)} \\ \scriptsize{0.692} \tiny{(0.006)} \\ \scriptsize{\underline{\textbf{0.711}}} \tiny{(0.006)} \\ \scriptsize{0.452} \tiny{(0.009)} \end{tabular} & \begin{tabular}{@{}c@{}} \scriptsize{0.833} \tiny{(0.010)} \\ \scriptsize{0.831} \tiny{(0.012)} \\ \scriptsize{\textbf{0.895}} \tiny{(0.009)} \\ \scriptsize{0.302} \tiny{(0.014)} \end{tabular} & \begin{tabular}{@{}c@{}} \scriptsize{\textbf{0.615}} \tiny{(0.012)} \\ \scriptsize{0.585} \tiny{(0.011)} \\ \scriptsize{0.571} \tiny{(0.007)} \\ \scriptsize{0.564} \tiny{(0.006)} \end{tabular} & \begin{tabular}{@{}c@{}} \scriptsize{0.740} \tiny{(0.004)} \\ \scriptsize{0.723} \tiny{(0.007)} \\ \scriptsize{\underline{\textbf{0.750}}} \tiny{(0.005)} \\ \scriptsize{0.492} \tiny{(0.010)} \end{tabular} & \begin{tabular}{@{}c@{}} \scriptsize{0.833} \tiny{(0.010)} \\ \scriptsize{0.831} \tiny{(0.012)} \\ \scriptsize{\textbf{0.895}} \tiny{(0.009)} \\ \scriptsize{0.302} \tiny{(0.014)} \end{tabular} & \begin{tabular}{@{}c@{}} \scriptsize{\underline{\textbf{0.670}}} \tiny{(0.010)} \\ \scriptsize{0.640} \tiny{(0.011)} \\ \scriptsize{0.640} \tiny{(0.007)} \\ \scriptsize{0.634} \tiny{(0.005)} \end{tabular} & \begin{tabular}{@{}c@{}} \scriptsize{\textbf{234.93}} \\ \scriptsize{584.77} \\ \scriptsize{337.80} \\ \scriptsize{2739.06} \end{tabular} \\
    \midrule
      \footnotesize{Token Cl.} & \begin{tabular}{@{}l@{}} \small{\texttt{distilbert-base-uncased}} \\ \small{\texttt{bert-base-uncased}} \\[1.7mm] \small{\texttt{distilroberta-base}} \\ \small{\texttt{roberta-base}} \end{tabular} & \begin{tabular}{@{}c@{}} \scriptsize{0.553} \tiny{(0.021)} \\ \scriptsize{0.546} \tiny{(0.017)} \\[1.7mm] \scriptsize{0.638} \tiny{(0.014)} \\ \scriptsize{\textbf{0.738}} \tiny{(0.010)} \end{tabular} & \begin{tabular}{@{}c@{}} \scriptsize{0.595} \tiny{(0.019)} \\ \scriptsize{0.587} \tiny{(0.019)} \\[1.7mm] \scriptsize{0.747} \tiny{(0.012)} \\ \scriptsize{\textbf{0.838}} \tiny{(0.018)} \end{tabular} & \begin{tabular}{@{}c@{}} \scriptsize{0.521} \tiny{(0.030)} \\ \scriptsize{0.516} \tiny{(0.022)} \\[1.7mm] \scriptsize{0.552} \tiny{(0.025)} \\ \scriptsize{\textbf{0.657}} \tiny{(0.022)} \end{tabular} & \begin{tabular}{@{}c@{}} \scriptsize{0.513} \tiny{(0.020)} \\ \scriptsize{0.506} \tiny{(0.015)} \\[1.7mm] \scriptsize{0.592} \tiny{(0.013)} \\ \scriptsize{\textbf{0.697}} \tiny{(0.008)} \end{tabular} & \begin{tabular}{@{}c@{}} \scriptsize{0.595} \tiny{(0.019)} \\ \scriptsize{0.587} \tiny{(0.019)} \\[1.7mm] \scriptsize{0.747} \tiny{(0.012)} \\ \scriptsize{\textbf{0.838}} \tiny{(0.018)} \end{tabular} & \begin{tabular}{@{}c@{}} \scriptsize{0.452} \tiny{(0.030)} \\ \scriptsize{0.448} \tiny{(0.021)} \\[1.7mm] \scriptsize{0.470} \tiny{(0.027)} \\ \scriptsize{\textbf{0.585}} \tiny{(0.023)} \end{tabular} & \begin{tabular}{@{}c@{}} \scriptsize{0.513} \tiny{(0.020)} \\ \scriptsize{0.506} \tiny{(0.015)} \\[1.7mm] \scriptsize{0.592} \tiny{(0.012)} \\ \scriptsize{\textbf{0.690}} \tiny{(0.008)} \end{tabular} & \begin{tabular}{@{}c@{}} \scriptsize{0.595} \tiny{(0.019)} \\ \scriptsize{0.587} \tiny{(0.019)} \\[1.7mm] \scriptsize{0.747} \tiny{(0.012)} \\ \scriptsize{\textbf{0.838}} \tiny{(0.018)} \end{tabular} & \begin{tabular}{@{}c@{}} \scriptsize{0.451} \tiny{(0.029)} \\ \scriptsize{0.446} \tiny{(0.021)} \\[1.7mm] \scriptsize{0.471} \tiny{(0.024)} \\ \scriptsize{\textbf{0.574}} \tiny{(0.022)} \end{tabular} & \begin{tabular}{@{}c@{}} \scriptsize{0.62} \\ \scriptsize{1.11} \\[1.7mm] \scriptsize{\underline{\textbf{0.57}}} \\ \scriptsize{1.23} \end{tabular} \\
    \midrule
      \footnotesize{Baselines} & \begin{tabular}{@{}l@{}} {\fontsize{7.5}{10}\selectfont No-DSD} \\ {\fontsize{7.5}{10}\selectfont Naive-DSD} \end{tabular} & \begin{tabular}{@{}c@{}} \scriptsize{0.429} \tiny{(0.020)} \\ \scriptsize{\textbf{0.455}} \tiny{(0.018)} \end{tabular} & \begin{tabular}{@{}c@{}} \scriptsize{\underline{\textbf{1.000}}} \tiny{(0.000)} \\ \scriptsize{0.003} \tiny{(0.002)} \end{tabular} & \begin{tabular}{@{}c@{}} \scriptsize{0.000} \tiny{(0.000)} \\ \scriptsize{\textbf{0.793}} \tiny{(0.014)} \end{tabular} & \begin{tabular}{@{}c@{}} \scriptsize{\textbf{0.429}} \tiny{(0.020)} \\ \scriptsize{0.266} \tiny{(0.011)} \end{tabular} & \begin{tabular}{@{}c@{}} \scriptsize{\underline{\textbf{1.000}}} \tiny{(0.000)} \\ \scriptsize{0.003} \tiny{(0.002)} \end{tabular} & \begin{tabular}{@{}c@{}} \scriptsize{0.000} \tiny{(0.000)} \\ \scriptsize{\textbf{0.462}} \tiny{(0.008)} \end{tabular} & \begin{tabular}{@{}c@{}} \scriptsize{\textbf{0.429}} \tiny{(0.020)} \\ \scriptsize{0.311} \tiny{(0.013)} \end{tabular} & \begin{tabular}{@{}c@{}} \scriptsize{\underline{\textbf{1.000}}} \tiny{(0.000)} \\ \scriptsize{0.003} \tiny{(0.002)} \end{tabular} & \begin{tabular}{@{}c@{}} \scriptsize{0.000} \tiny{(0.000)} \\ \scriptsize{\textbf{0.542}} \tiny{(0.010)} \end{tabular} & \begin{tabular}{@{}c@{}} \scriptsize{N/A} \\ \scriptsize{N/A} \end{tabular} \\
    \bottomrule
  \end{tabularx}
  \caption{Mean results achieved on the SSD by the various DSD methods, evaluated via 5-fold cross-validation. We include standard errors within parenthesis. For methods that require a threshold, its value is included in parenthesis next to the model's name. Best results from each method are shown in bold, while best results overall in a column are also underlined. The time column reflects the total time to run the entire evaluation (i.e., 5 folds). As already mentioned, for methods that rely on external APIs (e.g., LLM-DSD), times might fluctuate depending on rate limits, connection speeds, etc. The elevated time for the DeepSeek R1 model was due to the fact that the model was often formatting the output incorrectly, and therefore a lot of retries were needed.}\label{tab:ssd-results}
\end{sidewaystable}

\newpage

\vspace{1cm}
\section{Examples of Annotated Outputs}
\label{sec:appendix-annotated-output-examples}

\autoref{tab:output-examples} collects four different examples of annotations outputted by the different methods that we have presented, together with the reference annotation for comparison.

\hspace{0.5cm}

\newcolumntype{L}[1]{>{\raggedright\arraybackslash}p{#1}}
\renewcommand{\arraystretch}{2}
\small
\begin{longtable}{@{}L{3.7cm}L{3.7cm}p{3.15cm}L{4cm}@{}}
    \toprule
       \textbf{Sentence 1} & \textbf{Sentence 2} &  \textbf{Method} & \textbf{Annotation} \\
    \midrule
        \monospace{Following the Civil War, the plantation house was destroyed by hurricanes.} & \monospace{Before the Civil War, hurricanes destroyed the plantation house.} & Reference \newline Annotation & \monospace{\anndiss{Before the Civil War}, hurricanes destroyed the plantation house.} \\

        & & LIME \newline \texttt{all-MiniLM-L6-v2} \newline \scriptsize{$\alpha = \text{0.030}$} & \monospace{\anndiss{Before the} Civil \anndiss{War}, hurricanes destroyed the \anndiss{plantation house.}} \\

        & & SHAP \newline \texttt{all-MiniLM-L6-v2} \newline \scriptsize{$\alpha = \text{0.010}$} & \monospace{Before the Civil War\anndiss{,} hurricanes destroyed the plantation house\anndiss{.}} \\

        & & Embedding \newline \texttt{all-MiniLM-L6-v2} \newline \scriptsize{$\alpha = \text{0.010}$} & \monospace{Before the Civil War, hurricanes destroyed the plantation house.} \\

        & & Embedding \newline \texttt{text-embedding-3-large} \newline \scriptsize{$\alpha = \text{0.005}$} & \monospace{\anndiss{Before} the Civil War, hurricanes destroyed the plantation house.} \\

        & & LLM \newline \texttt{GPT-4o} & \monospace{Before \anndiss{the Civil War}, hurricanes destroyed the plantation house.} \\

        & & Token Classification \newline \texttt{roberta-base} & \monospace{\anndiss{Before the} Civil War, hurricanes destroyed the plantation house.} \\

        & & Baseline \newline \texttt{Naive-DSD} & \monospace{\anndiss{Before} the Civil War, hurricanes destroyed the plantation house.} \\

    \midrule

        \monospace{Retail version supports RAM disk size between 5MiB to 64GiB.} & \monospace{Enterprise version supports RAM disk size between 1GiB to 128GiB.} & Reference \newline Annotation & \monospace{\anndiss{Enterprise version} supports RAM disk size \anndiss{between 1GiB to 128GiB}.} \\

        & & LIME \newline \texttt{all-mpnet-base-v2} \newline \scriptsize{$\alpha = \text{0.015}$} & \monospace{Enterprise version supports RAM disk size between 1GiB to 128GiB.} \\

        & & SHAP \newline \texttt{all-mpnet-base-v2} \newline \scriptsize{$\alpha = \text{0.030}$} & \monospace{\anndiss{Enterprise} version supports RAM disk size between 1GiB \anndiss{to} 128GiB.} \\

        & & Embedding \newline \texttt{all-mpnet-base-v2} \newline \scriptsize{$\alpha = \text{0.006}$} & \monospace{\anndiss{Enterprise version} supports RAM disk size between 1GiB to 128GiB\anndiss{.}} \\

        & & Embedding \newline \texttt{text-embedding-004} \newline \scriptsize{$\alpha = \text{0.005}$} & \monospace{\anndiss{Enterprise version supports} RAM disk size \anndiss{between 1GiB to 128GiB.}} \\

        & & LLM \newline \texttt{Claude 3.5 Sonnet} & \monospace{\anndiss{Enterprise version} supports RAM disk size \anndiss{between 1GiB to 128GiB.}} \\

        & & Token Classification \newline \texttt{distilroberta-base} & \monospace{\anndiss{Enterprise version} supports RAM disk size between \anndiss{1GiB} to \anndiss{128GiB}.} \\

        & & Baseline \newline \texttt{Naive-DSD} & \monospace{\anndiss{Enterprise} version supports RAM disk size between \anndiss{1GiB} to \anndiss{128GiB}.} \\

    \midrule

        \monospace{China stock index futures close higher -- Dec. 4} & \monospace{China stock index futures close lower -- Jan. 24} & Reference \newline Annotation & \monospace{China stock index futures close \anndiss{lower} -- Jan. 24} \\

        & & LIME \newline \texttt{all-MiniLM-L6-v2} \newline \scriptsize{$\alpha = \text{0.010}$} & \monospace{China \anndiss{stock index} futures \anndiss{close lower} -- \anndiss{Jan}. \anndiss{24}} \\

        & & SHAP \newline \texttt{all-MiniLM-L6-v2} \newline \scriptsize{$\alpha = \text{0.005}$} & \monospace{China \anndiss{stock} index futures close lower \anndiss{-- Jan. 24}} \\

        & & Embedding \newline \texttt{all-MiniLM-L6-v2} \newline \scriptsize{$\alpha = \text{0.005}$} & \monospace{China stock index futures close lower -- \anndiss{Jan. 24}} \\

        & & Embedding \newline \texttt{text-embedding-3-large} \newline \scriptsize{$\alpha = \text{0.005}$} & \monospace{\anndiss{China stock index futures close lower -- Jan. 24}} \\

        & & LLM \newline \texttt{Llama 3.3 \newline 70B Instruct} & \monospace{China stock index futures \anndiss{close lower} -- \anndiss{Jan. 24}} \\

        & & Token Classification \newline \texttt{bert-base-uncased} & \monospace{China stock index futures close \anndiss{lower --} \anndiss{Jan. 24}} \\

        & & Baseline \newline \texttt{Naive-DSD} & \monospace{China stock index futures close \anndiss{lower} -- \anndiss{Jan. 24}} \\
    \bottomrule
    \captionsetup{width=1\textwidth}
    \caption{Examples of annotated outputs. The first two sentence pairs are coming from the SSD, while the third proceeds from the SemEval-2016 Task 2 data. In the latter, it can be seen that the differing dates ``Dec. 4'' and ``Jan. 24'' have not been included in the reference annotations (they were annotated as \textit{similar}). This is one of the caveats with the SemEval-2016 Task dataset: there are no unambiguous annotations of dissimilarity beyond the \textit{opposite} spans.}
    \label{tab:output-examples}
\end{longtable}
\normalsize

\section{Embedding-DSD Algorithm}
\label{sec:appendix-embedding-dsd-algorithm}

\autoref{alg:token-embedding-dsd} and \autoref{alg:calculate-replacements} describe the steps involved in the Embedding-DSD method. \autoref{alg:token-embedding-dsd} returns a map (dictionary) specifying the dissimilarity score for each of the unigrams in the second sentence. After that, it still remains applying the threshold over these scores in order to determine which unigrams will be included within a dissimilar span.\\[1cm]

\begin{algorithm}
	\setstretch{1.35}
	\caption{Embedding-DSD.}\label{alg:token-embedding-dsd}
	\begin{algorithmic}[1]
		\Procedure{Embedding-DSD}{$sentence1, sentence2$}
		\State $baseSimilarity \gets \text{cosSim}(sentence1, sentence2)$
		\State $replacements \gets \text{getReplacements}(sentence1, sentence2)$
        \State $similarityGains \gets \{\,\}$
        \For{$indices, \, repl \in replacements$} \Comment{Replaced unigram indices, and resulting string}
            \State $gain \gets \text{cosSim}(sentence1, repl) - baseSimilarity$ \Comment{Similarity gain yielded by the replacement}
			\State $similarityGains[indices] \gets gain$
		\EndFor
        \State $unigrams \gets \text{getUnigrams}(sentence2)$
        \State $unigramDiff \gets \{\,\}$ \Comment{Map from unigram index to array of gains}
        \For{$uniIndex \gets 0, \, \text{len(}unigrams) - 1$} \Comment{Assign \textit{n}-gram gains to individual unigrams}
        \vspace{-0.5cm}
            \For{$gainIndices, \, gain \in similarityGains$}
            \vspace{-0.5cm}
                \If{$uniIndex \in gainIndices $}
                    \vspace{-0.1cm}
                    \State $unigramDiff[uniIndex].\text{insert}(gain)$
                \EndIf
            \EndFor
        \EndFor
        \For{$uniIndex \in unigramDiff$} \Comment{Aggregate gains}
            \State $unigramDiff[uniIndex] \gets \text{aggregateGains}(unigramDiff[uniIndex])$
        \EndFor
		\State \textbf{return} $unigramDiff$
		\State \hspace{-0.5cm}\textbf{end procedure}
		\EndProcedure
	\end{algorithmic}
\end{algorithm}

\begin{algorithm}
	\setstretch{1.35}
	\caption{Calculate all possible \textit{n}-gram replacements.}\label{alg:calculate-replacements}
	\begin{algorithmic}[1]
		\Procedure{GetReplacements}{$sentence1, sentence2$}
		\State $sent2Unigrams \gets \text{getUnigrams}(sentence2)$
        \State $sent2Replacements \gets \{\,\}$ \Comment{Map from replaced indices to resultant string}
        \For{$nGramSize \gets 1, \, \text{len}(sent2Unigrams)$} \Comment{Consider all possible \textit{n}-gram sizes}
            \State $sent1NGrams \gets \text{getNGrams}(sentence1, nGramSize)$
            \vspace{-0.25cm}
            \For{$nGram \in sent1NGrams$} \Comment{Generate replacements for each retrieved \textit{n}-gram}
                \vspace{-0.45cm}
                \For{$i \gets 0, \, \text{len}(sent2Unigrams)  - \text{len}(nGram) + 1$} \Comment{Positions that allow a replacement}
                    \State $replacedIndices \gets [\,]$
                    \State $replaced \gets sent2Unigrams.\text{copy()}$
                    \For{$j \gets 0, \, \text{len}(nGram)$} \Comment{Replace the unigrams present in the \textit{n}-gram}
                        \State $replacedIndices.\text{insert}(i + j)$
                        \State $replaced[i + j] = nGram[j]$
                    \EndFor
                    \State $sent2Replacements[replacedIndices] \gets replaced.\text{toString()}$
                \EndFor
            \EndFor
        \EndFor
		\State \textbf{return} $sent2Replacements$
		\State \hspace{-0.5cm}\textbf{end procedure}
		\EndProcedure
	\end{algorithmic}
\end{algorithm}
\setstretch{1}

\vspace{0.5cm}
\section{LLM Prompts}
\label{sec:appendix-llm-prompts}

Here, we include the prompts that we used for 1) generating the differing spans in the SSD, and 2) evaluating different LLMs on the SSD.

\vspace{0.2cm}
\subsection{Prompt used for the SSD creation}

As mentioned earlier, we initially tried to prompt the LLM to generate both the modified spans, as well as their labels (i.e., 0 if the modified span pair was semantically dissimilar, or 1 if it was equivalent in meaning). However, since the labels were often incorrectly annotated, we resorted to labeling them ourselves. To get the annotated sentence, we used the following prompt:

\vspace{0.5cm}

\begin{prompt}
    I will give you some sentences below, each of them separated by a newline. Your task is to modify spans of text. The modified spans can either the same meaning as the original, or an opposite meaning.  Please, annotate your spans in both the original and the modified sentence with the characters \textasciigrave\{\{\textasciigrave for the beginning of a span, and \textasciigrave\}\}\textasciigrave for its ending. Also, try to replace spans containing more than one word. For example:

    \vspace{0.2cm}

    Input 1:
    
    \q{Thinking about it, the issue might be more critical than what we first thought.}

    \vspace{0.2cm}

    Output 1:

    \q{Thinking about it, the issue \{\{might be more critical\}\} than what we \{\{first\}\} thought.} => \q{Thinking about it, the issue \{\{is less relevant\}\} than what we \{\{initially\}\} thought.}

    \vspace{0.5cm}
    
    Input 2:

    \q{She gazed out of the window, contemplating the distant city lights twinkling in the evening darkness.}

    \vspace{0.2cm}

    Output 2:

    \q{She gazed \{\{out of the window\}\}, \{\{contemplating\}\} the distant city lights twinkling in the evening \{\{darkness\}\}.} => \q{She gazed \{\{inside her heart\}\}, \{\{pondering\}\} the distant city lights twinkling in the evening \{\{obscurity\}\}.}

    \vspace{0.5cm}
    
    Here are the sentences to annotate. Give me only the output as provided in the example above, and don't leave blank lines between outputs:

    \vspace{0.2cm}

    \q{It has been reported from a number of states.}

    \q{It has a W10 loading gauge but W9 rolling stock is excluded.}

    [...]

    \q{Timberlacing includes finely webbed Dhajji.}

    \q{Other programs exclude the famous CUNY College Now and GEAR UP.}

    \vspace{0.2cm}
    
    Don't forget to include spans that have equivalent meaning.
\end{prompt}

In our prompts, we included around 35 sentences (above omitted with \q{[...]} for brevity). The last instruction was added due to the model’s tendency to output uniquely dissimilar spans. Some example outputs for the previous input would be:

\begin{prompt}
    \q{It has \{\{been reported from\}\} a number of \{\{states\}\}." => "It has \{\{come in from\}\} a number of \{\{regions\}\}.}
    
    \q{It has a \{\{W10 loading gauge\}\} but \{\{W9 rolling stock\}\} is excluded." => "It has a \{\{high capacity gauge\}\} but \{\{smaller stock\}\} is excluded.}

    [...]
    
    \q{\{\{Timberlacing includes\}\} finely webbed \{\{Dhajji\}\}." => "\{\{Wooden reinforcements feature\}\} finely webbed \{\{timber frames\}\}.}
    
    \q{Other programs \{\{exclude the famous\}\} CUNY College Now and \{\{GEAR UP\}\}." => "Other programs \{\{leave out the well-known\}\} CUNY College Now and \{\{the GEAR UP initiative\}\}.}
\end{prompt}

\vspace{0.2cm}
\subsection{Prompt used in LLM-DSD}

In order to evaluate LLMs on the SSD, we adopt an in-context learning setting, providing within the prompt four examples with the expected inputs and outputs.

First, the following message is passed to the model as \textit{system prompt}:

\begin{prompt}
    You are an NLP model able to detect differences in meaning in textual pairs. More concretely, given a premise and a hypothesis, you are able to compare them and annotate in the hypothesis the spans that are differing in meaning to the information included in the premise.

    Here is an example of an input and the expected output:

    \begin{lstlisting}
# INPUT 1

```json
{
  "premise": "There was international outrage for the decision.",
  "hypothesis": "There was no reaction to the decision."
}
```\end{lstlisting}

    \begin{lstlisting}
# OUTPUT 1

```
There was {{no reaction}} to the decision.
```\end{lstlisting}

    \noindent As you can see, the inputs are formatted as a JSON blob containing the premise and hypothesis. The response is a code block containing the  hypothesis with the differing spans enclosed within the markers "\{\{" and "\}\}". Note that without these markers, both the input hypothesis and the annotated hypothesis are identical.
    
    \noindent Here is another example:

    \begin{lstlisting}
# INPUT 2

```json
{
  "premise": "Microorganisms are too small to be seen by the naked eye.",
  "hypothesis": "Microorganisms have considerable size and can be seen with your eyes."
}
```\end{lstlisting}

    \begin{lstlisting}
# OUTPUT 2

```
Microorganisms {{have considerable size}} and {{can be seen with your eyes}}.
```\end{lstlisting}
    
    \noindent On the other hand, here goes an example with a incorrect output, since the annotated hypothesis includes words not present in the input hypothesis:

    \begin{lstlisting}
# INPUT 3

```json
{
  "premise": "It is much warmer here than it used to be.", 
  "hypothesis": "It is way colder here than it used to be."
}
```\end{lstlisting}

    \begin{lstlisting}
# (INCORRECT) OUTPUT 3

```
I believe it is {{way colder}} here than it used to be.
```\end{lstlisting}
    
    \noindent Let me show you one last example of an erroneous output, in this case because the annotation appears at the end and not within the hypothesis: 

    \begin{lstlisting}
# INPUT 4

```json
{
  "premise": "The deputy was urged to provide an immediate apology for his controversial comments.", 
  "hypothesis": "The deputy was urged to resign for his controversial comments."
}
```\end{lstlisting}

    \begin{lstlisting}
# (INCORRECT) OUTPUT 4

```
The deputy was urged to resign for his controversial comments. {{resign}}
```\end{lstlisting}
\end{prompt}

Then, the specific instance to be annotated was passed as a \textit{user message}:

\begin{prompt}
    \noindent You are now given the following JSON input:

\begin{lstlisting}
```json
{
  "premise": "He worked as a curator, editor and bibliographer.",
  "hypothesis": "He worked as a gardener, writer, and librarian."
}
```
\end{lstlisting}
    \noindent Please, provide the annotated hypothesis using the start marker "\{\{" and the end marker "\}\}". Enclose the annotated hypothesis within a code block using "\textasciigrave \textasciigrave \textasciigrave" so I can easily identify it. Please, reason your answer.
\end{prompt}

\section{Instructions for Annotators}
\label{sec:instructions-annotators}

In this section, we present the specific instructions that were provided to the annotators, which were intended to guide them in the validation process and ensure the accuracy and correctness of the SSD.

\begin{quote}
    \noindent\rule{\linewidth}{0.5pt}

    \noindent \textbf{Span Similarity Dataset – Annotation Instructions} \\[-0.3cm]
    
    \noindent The data to be annotated is provided as a \texttt{.tsv} file called \texttt{annotate.tsv}. In each line, there are two sentences separated by a tab character (\texttt{\textbackslash t}). These sentences are very similar, except for certain word spans. Here we define a ``span'' as a group of contiguous words. In some cases, the differing spans are equivalent in meaning. In others, they have different meaning. Your task is to annotate the differing spans and determine whether they have equivalent meaning or not. More specifically:
    
    \begin{enumerate}
        \item Identify the differing spans and enclose them within double curly braces, i.e., \texttt{\{\{} to signal the beginning of a span, and \texttt{\}\}} to signal its end. Try to annotate the spans in such way that the spans have enough context on its own (e.g., in example 2, ``food'' and ``portions'' are included in spans).
        \begin{enumerate}
            \item Example:
                \begin{itemize}
                    \item Sentence 1: \texttt{I'm not a bad talker either.}
                    \item Sentence 2: \texttt{I'm not an eloquent speaker either.}
                    \item Annotated Sentence 1: \texttt{I'm not \{\{a bad talker\}\} either.}
                    \item Annotated Sentence 2: \texttt{I'm not \{\{an eloquent speaker\}\} either.}
                \end{itemize}
            \item Example:
                \begin{itemize}
                    \item Sentence 1: \texttt{Food is way overpriced and portions are very small.}
                    \item Sentence 2: \texttt{Food is reasonably priced and portions are generous.}
                    \item Annotated Sentence 1: \texttt{\{\{Food is way overpriced\}\} and \{\{portions are very small\}\}.}
                    \item Annotated Sentence 2: \texttt{\{\{Food is reasonably priced\}\} and \{\{portions are generous\}\}.}
                \end{itemize}
        \end{enumerate}
    
        \item Add a space to the end of the line. Then write a \texttt{1} if the annotated span is equivalent, or a \texttt{0} if it is dissimilar. In case there are several spans, separate the numbers with a comma (leave no space in between the numbers), e.g., \texttt{0,1}. Span pairs will always appear in the same order in both sentences.
        \begin{enumerate}
            \item Example:
                \begin{itemize}
                    \item Original Line: \\ \texttt{I'm not a bad talker either. I'm not an eloquent speaker either.}
                    \item Annotated Line: \\ \texttt{I'm not \{\{a bad talker\}\} either. I'm not \{\{an eloquent speaker\}\} either. 0}
                \end{itemize}
            \item Example:
                \begin{itemize}
                    \item Original Line: \\ \texttt{Food is way overpriced and portions are very small. Food is reasonably priced and portions are generous.}
                    \item Annotated Line: \\ \texttt{\{\{Food is way overpriced\}\} and \{\{portions are very small\}\}. \{\{Food is reasonably priced\}\} and \{\{portions are generous\}\}. 0,0}
                \end{itemize}
        \end{enumerate}
    \end{enumerate}
    
    You can find some examples of the resultant annotation in \textit{examples.tsv}. Please, \textbf{do not use any assistance from a Large Language Model (LLM)} or any other automatic annotation methods, since the goal is to evaluate human performance on the task. Save your annotations as \texttt{annotated\_{your-first-name}.tsv}. Thank you!

    \noindent\rule{\linewidth}{0.5pt}
\end{quote}

\section{Licenses of the Resources Employed}
\label{sec:resource-licenses}

Reporting the licenses of resources used is essential for ensuring transparency and reproducibility, allowing others to verify and build upon previous work without legal ambiguities. In \autoref{tab:dataset-licenses}, we collect the licenses of the datasets we used. \autoref{tab:model-licenses} is an analogous table collecting the model licenses.

\vspace{0.5cm}

\renewcommand{\arraystretch}{1.7}
\begin{table}[h]
    \centering
    \begin{tabular}{@{}lll@{}}
        \toprule
        \textbf{Dataset} & \textbf{License} & \textbf{Homepage} \\
        \midrule
        Span Similarity Dataset (SSD) & CC BY-SA 4.0 & \url{https://dmlls.github.io/dissimilar-span-detection} \\
        SemEval-2016 Task 2 Dataset & CC BY-SA 4.0 & \url{https://alt.qcri.org/semeval2016/task2} \\
        \bottomrule
    \end{tabular}
    \caption{Licenses of the datasets we used in our work.}
    \label{tab:dataset-licenses}
\end{table}

\vspace{0.5cm}

\renewcommand{\arraystretch}{1.4}
\begin{longtable}{@{}lp{3.5cm}p{6.1cm}@{}}
    \toprule
    \textbf{Model} & \textbf{License} & \textbf{Homepage} \\
    \midrule
    \endhead

    \texttt{all-MiniLM-L6-v2} & Apache License 2.0 & \href{https://huggingface.co/sentence-transformers/all-MiniLM-L6-v2}{\texttt{https://huggingface.co/sentence-}} \href{https://huggingface.co/sentence-transformers/all-MiniLM-L6-v2}{\texttt{transformers/all-MiniLM-L6-v2}} \\
    \texttt{all-mpnet-base-v2} & Apache License 2.0 & \href{https://huggingface.co/sentence-transformers/all-mpnet-base-v2}{\texttt{https://huggingface.co/sentence-}} \href{https://huggingface.co/sentence-transformers/all-mpnet-base-v2}{\texttt{transformers/all-mpnet-base-v2}} \\
    \texttt{text-embedding-3-large} & Proprietary & \url{https://platform.openai.com/docs/guides/embeddings/#embedding-models} \\
    \texttt{text-embedding-004} & Proprietary & \url{https://ai.google.dev/gemini-api/docs/embeddings} \\
    \texttt{GPT-4o} & Proprietary & \url{https://openai.com/index/hello-gpt-4o} \\
    \texttt{Llama 3.3 70B Instruct} & Llama 3.3 Community License Agreement & \url{https://www.llama.com/docs/model-cards-and-prompt-formats/llama3_3/} \\
    \texttt{Claude 3.5 Sonnet} & Proprietary & \url{https://www.anthropic.com/claude/sonnet} \\
    \texttt{DeepSeek R1} & MIT & \url{https://github.com/deepseek-ai/DeepSeek-R1} \\
    \texttt{bert-base-uncased} & Apache License 2.0 & \url{https://huggingface.co/google-bert/bert-base-uncased} \\
    \texttt{roberta-base} & MIT & \url{https://huggingface.co/FacebookAI/roberta-base} \\
    \texttt{distilbert-base-uncased} & Apache License 2.0 & \url{https://huggingface.co/distilbert/distilbert-base-uncased} \\
    \texttt{distilroberta-base} & Apache License 2.0 & \url{https://huggingface.co/distilbert/distilroberta-base} \\
    \bottomrule
    \caption{Licenses of the models we used in our work.}
    \label{tab:model-licenses}
\end{longtable}

\end{document}